# The scope for AI-augmented interpretation of building blueprints in commercial and industrial property insurance.

A vision of future assessment of property fire and explosion risks.

Long Chen[1], Mao Ye[2], Alistair Milne[3], John Hillier[4] and Frances Oglesby[5]

Loughborough University

## Abstract

This report, commissioned by the WTW research network, investigates the use of AI in property risk assessment. It (i) reviews existing work on risk assessment in commercial and industrial properties and automated information extraction from building blueprints; and (ii) presents an exploratory 'proof-of concept-solution' exploring the feasibility of using machine learning for the automated extraction of information from building blueprints to support insurance risk assessment.

## Table of Contents

[1] School of Architecture, Building and Civil Engineering; email: L.Chen3@lboro.ac.uk
[2] Department of Computer Science, School of Science; email: M.Ye@lboro.ac.uk
[3] School of Business and Economics; email: a.k.l.milne@lboro.ac.uk
[4] Department of Geography, School of Social Sciences and Humanities; email: J.Hillier@lboro.ac.uk
[5] School of Business and Economics; email: f.a.oglesby-16@student.lboro.ac.uk

# Exec summary

Technology has had a transformative impact on many areas of insurance. One sub-market still urrently still largely reliant on manual processes is for the insurance of industrial properties, especially insurance against the critical risks of fire and explosion. Is this an area of insurance ripe for digital assistance? If so, what will be the impact on clients, brokers and underwriters?

This paper reports an initial investigation into the use of machine learning tools, commonly referred to as 'AI', for blueprint interpretation. These blueprints, particularly for older properties, are a key source of information for risk engineer's but typically only available on paper or as pdf documents. Such digitisation could reduce much of the routine work of risk engineers, allowing them to focus on where they add most value, using their experience and skills to make the professional judgements of potential loss on which the terms of insurance policies are based. Digitisation could also lead to more effective and standardised sharing of information on exposures and vulnerability, in turn helping address some of the substantial frictions that increase premiums and limit available coverage.

This work is a pilot study using publicly available residential property blueprints and focussing as a final output on calculation of room area. The investigation has required harnessing and integrating a number of currently available technologies. It shows that it is possible to combine and interpret graphical, locational and textual blueprint information. It has led to a good understanding of what is required for these methods to be used at scale for property risk assessment, but also highlighted the challenges of implementation.

A principal insight is that there is no inherent barrier to information being automatically extracted from industrial and commercial property blueprints. This is a challenge that can be met with sufficient access to blueprints and time spent on developing the processing and training the software. While outcome will never be 100% accurate, high levels of accuracy can be achieved provided the processed blueprints are not totally variant from those used for the software development. The key advantage is processing speed. These tools can provide useful practical support to risk engineers, relieving them of much routine work and leaving them with the easier task of reviewing the outputs of automated blued print interpretation.

The technical contribution has been the successful application of a selected combination of machine learning tools. These are not themselves new, they are well established in a variety of computer vision applications including to blueprints and architectural drawings developed over the past decade (reviewed in an Appendix). The added value is integrating the tools into a single integrated 'end to end' process, that begins with blueprints as inputs, extracts information in a useable form and produces analytical output.

The paper also offers considers the further evolution of business process, building on the foundation of automated blueprint interpretation, leading to wider economic benefits. Digitisation of this kind can further support standardisation of business processes and enhance consistency of approach across individuals and firms. If practitioners adopt common approaches and these are accepted at national and industry level, this can support greater comparability of risk assessment and thus increased efficiency of risk transfer as well as reducing cost in risk assessment. Digitisation and adoption of more standardised methods can further support the development of tools of knowledge management for risk engineering, ensuring that skills and understanding are understood, shared and disseminated as widely as possible.

# 1. Introduction

This report presents the outcome of a four-month investigation into the feasibility of using automated information extraction from building blueprints to support assessments by risk engineers of insurance risk in industrial and commercial properties.

The motivation for this work is two-fold, using technology to promote efficiency and consistency in risk assessment. Potential for increased efficiency comes from the observation that the experienced risk engineers, working on behalf of clients to assess their requirements for property insurance, do a great deal of manual extraction of information from building blueprints. The question then arises whether it would be possible to develop solutions to automate the extraction of at least some information extraction from blueprints. Ideally, solutions would provide risk engineers with convenient automated summaries of key risk metrics, such as the potential spread of a fire or explosion event from different locations or the effectiveness of active fire protection technologies such as sprinkler systems. Such solutions could free up a substantial amount of time, allowing risk engineers to focus more on applying their professional judgement to overall risk assessment.

Potential for increased consistency comes from the observation that it is currently possible for risk engineers to diverge in their views of a particular property. Automation of key risk metrics could provide a benchmark, within and between firms, that risk engineers can then use as a standardized starting place for their view of risk, aiding consistency. To this, there is the added potential advantage that existing learning within teams can be captured and systematised, reducing their sensitivity to individuals moving on.

We emphasise that this is still a preliminary analysis. After we were given a short introduction to the work of WTW risk engineers in industrial and commercial property risk assessment, we focused on developing a 'proof of concept' solution, combining a variety of computational methods for effective information extraction and analysis from public domain blueprints. This was done for residential rather than industrial and commercial buildings as the task is simpler. We supported the computational work with a brief review of relevant literature, as a check that the processes that were carried out in our proof of concept, if carried forward to blueprints for industrial and commercial properties would provide information that could be useful support to practising risk engineers.

While still preliminary, our findings are encouraging. They suggest that there is no inherent barrier to using machine learning, also known as artificial intelligence or AI, solutions for automated extraction of information of value to risk engineers from blueprints for industrial and commercial properties. It is also clear that developing these solutions will require a quite extensive exercise in software development, applied to a wide variety of properties and property blueprints. It seems worthwhile to take this a stage further with a detailed engagement with potential users, professional risk engineers, and the various types of data and data requirements, including from blueprints, that they use in their day-to-day work.

Finally – building on related analysis conducted over the past three years at Loughborough University in the TECHNGI project on technology innovation in insurance – this report also briefly considers the broader challenges for adoption of AI and other digital technologies in industrial and commercial property insurance.

The report is set out as follows. Section 2 is an overview of relevant research work, both in property insurance risk and on automated extraction of blueprint information. Section 3 summarises our proof-of-concept work. Section 4 concludes with a view on potential future avenues of work.

There are three supporting appendices: Appendix A on the property insurance risk literature; Appendix B on the existing research on automated blueprint interpretation. Appendix C is the detailed presentation of the methods and output of our integrated proof of concept solution.

# 2. Existing research

This section is an overview of relevant existing research on the two topics integral to understanding the feasibility of automated blueprint information extraction in support of this risk assessment. First, the assessment of insurance risks in industrial and commercial property (Appendix A), and second on the automated analysis of blueprints (Appendix B).

## Risk assessment in industrial and commercial properties

Analysis by the University of Southern Denmark has reviewed professional practice in the assessment of risks for industrial and commercial property insurance. To complement that study, we conducted a short further review of existing research literature on risk assessment in property insurance with a focus on fire risk (detail in Appendix A).

As a conceptual framework, this review employs the standard accepted separation of overall risk of financial loss into three elements: the probability of the hazard event $P$, the exposure to potential loss $E$ and the vulnerability $V$ representing the actual loss experienced allowing for factors such as fire protections. In the current context, blueprint analysis is taken as central to assessing both how the layout of property affects the potential loss $E$ from e.g., a fire and vulnerability $V$ after allowing for protections such as firewalls and sprinkler systems. The main findings are as follows.

While there is only a limited literature, what there is confirms our expectation that automated blueprint assessment can be a useful supporting tool for insurance risk analysis of industrial and commercial properties, provided that it

(i) successfully identifies the layout of buildings, both internally and in relation to each other;
(ii) recognises and locates key relevant protections such as fire doors and fire walls
(iii) recognises and locates the presence of active protections, notably sprinkler systems

## Automated analysis of blueprints

We have also conducted a review of existing literature on automated blueprint analysis (reported in full in Appendix B). We here summarise the key points.

A building's blueprint is a fundamental of both property design and construction (Architecture, Engineering and Construction AEC) and property management (Facility Management FM). More recently-constructed buildings can have blueprints in electronic formats, generated by Computer-aided Design (CAD), Computer-aided Engineering (CAE) and Computer-Aided Manufacturing (CAM) files and incorporating digital information. This supports the automated digital and contextualized interpretation of buildings.

However, for most existing buildings blueprints are either pdfs held as computer files, or physical copies held in paper filing systems (see Fig. 1 for examples). These are limited to two dimensions and unlike CAD based electronic blueprints, do not incorporate digital information. The use of digital tools such as AI can reduce human effort involved in ensuring that the translation of building information into digital form is error-free (Rica, Moreno-García, Álvarez, & Serratosa, 2020)

The focus of this project has been on automated extraction of information from these basic blueprints, in order to develop as rich as possible set of supporting digital data that can in turn be used for supporting insurance analysis.

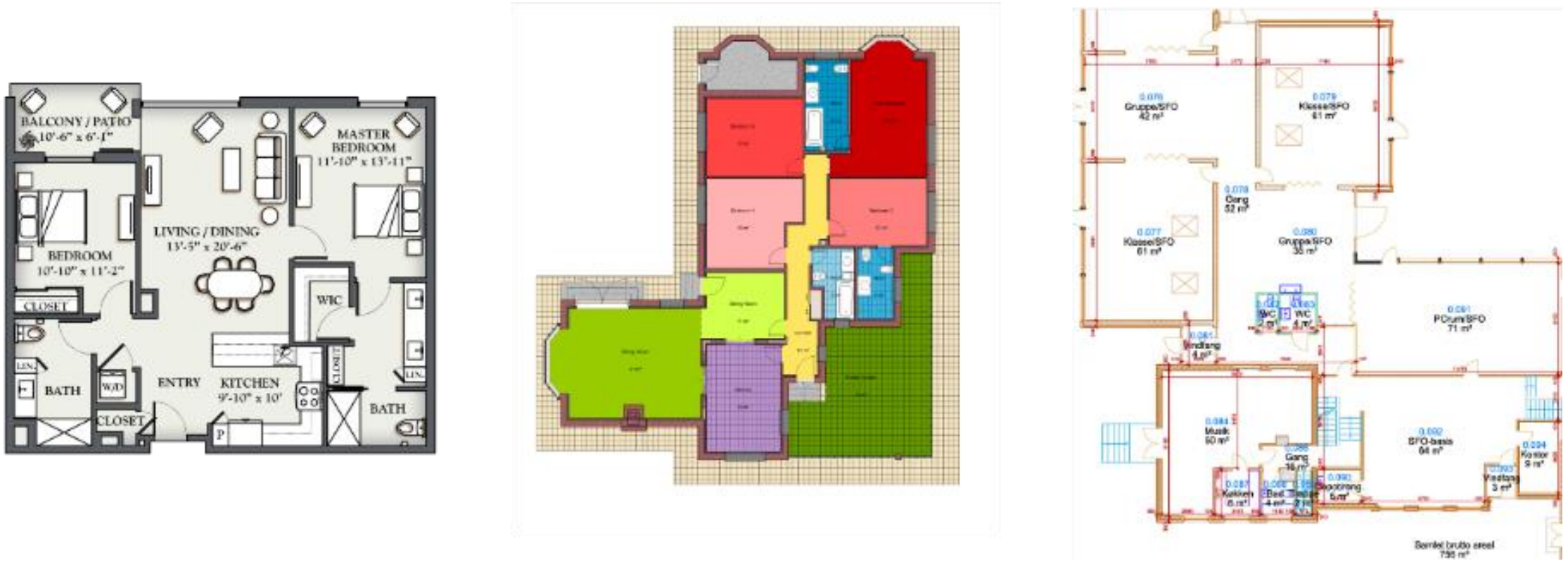

Fig. 1. Examples of various formats of blueprints

While there are no publications on previous application of machine learning or other AI technologies to building blueprints for insurance risk analysis, there is existing research on automated blueprint interpretation for architectural applications. This prior literature is summarised in Appendix B. Some early research from the 1990s developed traditional rule based computerised algorithms for ‘segmentation’ i.e., identifying and extracting location information for perimeter walls and interior dividing walls between rooms. Subsequent contributions implemented ‘raster to vector’ (R2V) transformations so that room boundaries can be captured in a simple and efficient way, represented as vectors rather than a constellation of pixels. A further application has been converting these into 3D models, in effect recreating some of the richer information incorporated into CAD modelling.

The most recent research identifies Artificial Intelligence (AI) including Computer Vision (CV), Optical Character Recognition (OCR) technologies as promising to support blueprint interpretation and analysis. Details are provided in Appendix B. One challenge is object detection and tracking which, illustratively, have been implemented in the construction industry with AI algorithms. Numerous AI approaches (e.g., Support Vector Machine (SVM), Convolution Neural Network (CNN), Deep Neural Network (DNN), Generative Adversarial Network (GAN) etc.) have been applied to a variety of data and processing issues. This object recognition can employ either “one stage” neural network methods which identify bounding boxes for the various structures and objects, or “two stage” approach in which a second refinement stage is added to confirm accuracy of object identification from their “convolution” i.e., relationship to each other.

In addition to object detection, Optical Character Recognition (OCR) technologies using Convolution Neural Network (CNN) are widely applied to the text extraction, such as the textual information relating to room categories (i.e., usage or type) and room size. These types of information can be straightforwardly extracted and output with OCR technologies.

While promising for application to insurance risk assessment, these previous investigations also reveal challenges. While recognising semantic information in blueprints is usually straightforward for construction experts, automatically interpreting and processing blueprints and recognising layout and semantics is not straightforward (Zeng, Li, Yu, & Fu, 2019). One example of the challenges that arise is that blueprints are produced according to varying design guidelines and drawing styles. **Error! Reference source not found.** gives three examples of the varying formats of blueprints and variation in notation, colour and font. The variety of blueprints is a substantial challenge for automated blueprints interpretation.

# 3. Initial work on AI applied to blueprints

This section is a summary of the initial proof-of-concept analysis of the potential for using AI for automated information extraction from building blueprints. This work was conducted using easily available public domain residential building blueprints. A fuller description is provided in Appendix C.

As shown in Fig. 2, the blueprint interpretation method employed is in two stages, first macro interpretation and then micro interpretation. The first stage is conducted with a focus on identification of primary outlines forming the basis of the building's structure. The processing elements include contour detection, main-wall detection, layout detection, area detection, colour detection (valuable if objects are colour coded) and so on. The second stage captures details, such as detection objects-of-interest (e.g., window, door, fire sprinklers, etc.) and textual information (e.g., room function, room position, room area, etc.).

For detecting objects of interest, the technologies include geometric recognition, and template matching Optical character recognition (OCR) is typically used for the text extraction.

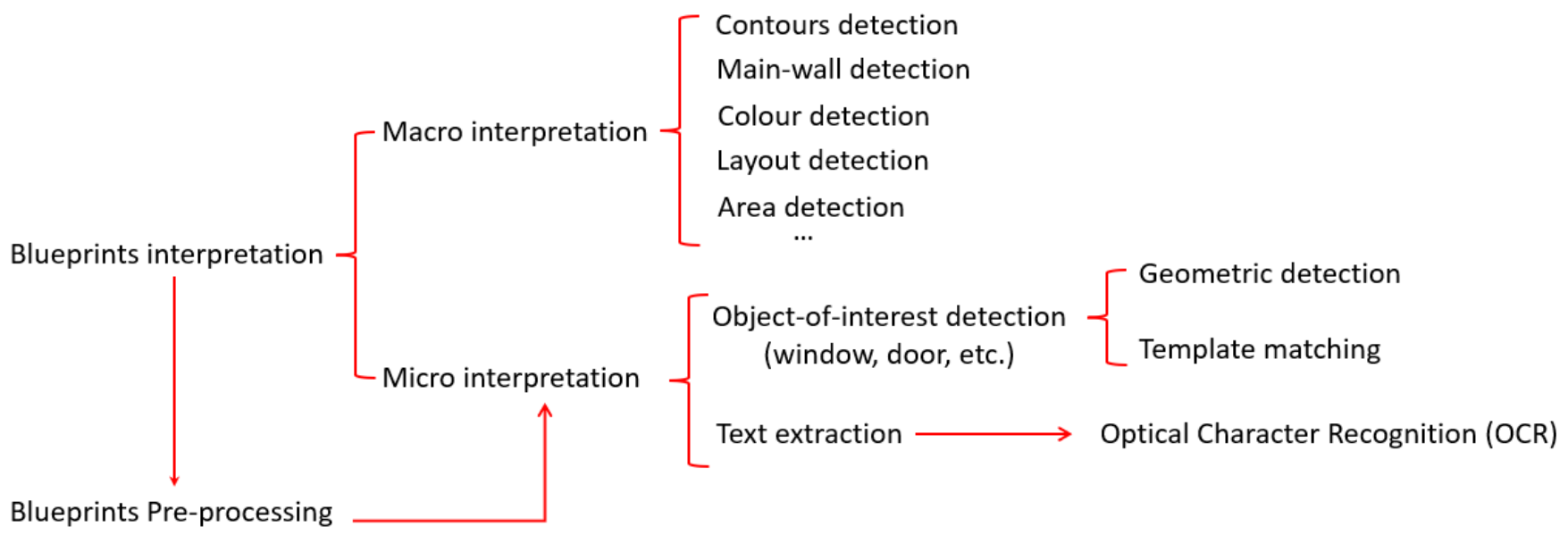

Fig. 2. Classification of blueprints interpretation

In conducting the blueprint interpretation in this pilot study, four stages of information extraction were proposed: blueprint pre-processing (to standardise format), capture of textual and structural information, recognition of graphical objects, and final the calibration of this information and integration of it into a coherent data structure. These are illustrated in Fig. 3.

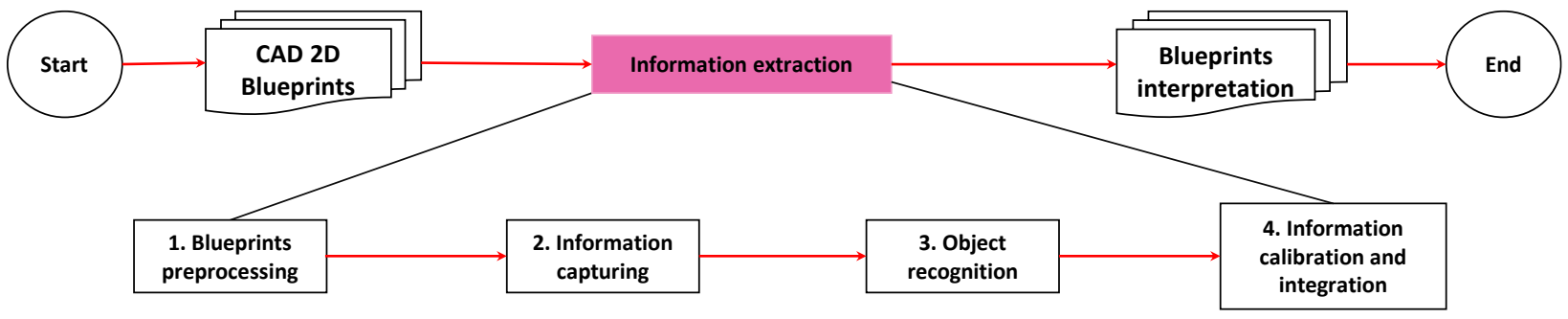

Fig. 3. The four stages of blueprints interpretation

In the proof-of-concept work, how these stage were enacted are outlined below, with more details are provided in Appendix C.

- *Blueprint pre-processing* was itself applied in several stages: i) resizing of blueprints, then ii) using statistical methods to filter out noise, followed by iii) binarization (conversion to two tone images) and colour inversion to further highlight key features and finally iv) 'morphological' operations that smooth out gaps and remove what appears to be extraneous information.
- *Information capturing* consists of both 'segmentation' (identifying room and building boundaries along with other structures) and 'optical character recognition' (reading of textual information).

- *Object recognition* is then carried out by matching structures, such as doors and windows, to templates, using various metrics of goodness of fit. Here it is more effective to employ multiple object matching, in which several similar objects in the same blueprint (e.g., multiple doors or multiple windows) are matched at the same time, reducing errors associated with single object matching. Similarly, textual information, initially extracted as isolated characters, can be recognised by reference to standard descriptors (in this residential case, descriptors such as kitchen or bedroom).
- *Information calibration and integration*. For this proof-of-concept, we have focused on the specific task of calculation of room areas, using the segmentation to compute relative room sizes and then relating this to information on lengths and on areas extracted using optical character recognition OCR, providing an independent verification of accuracy.

The proof-of-concept work provides an integrated solution for carrying out all four of these steps, demonstrating success in extracting of information from a variety of residential building blueprints, in identification of objects and in calculation of room areas.

Using the blueprints provided by Willis as an example (shown in Fig. 4), the original blueprints (left in Fig.4) have been pre-processed to reduce the background noise and image sizes using filtering algorithms. Then the walls (including interior and exterior walls) have been detected, based on which the whole blueprint is further segmented, and each component is coloured and numbered (lower middle in Fig.4). The pixels of scale detected in the original blueprint are further mapped to the real size, where the size of each room can be calculated out (lower right in Fig.4).

Given the rich textual information of blueprints, the embedded information can be further extracted and interpreted based on OCR techniques. This information can include room number, room function, room size, etc. and is marked out in the blueprints (upper middle in Fig.4). The marked information is finally extracted and stored in a separate text file, where the position information is marked in green, the room function in red, and room size in blue (upper right in Fig.4).

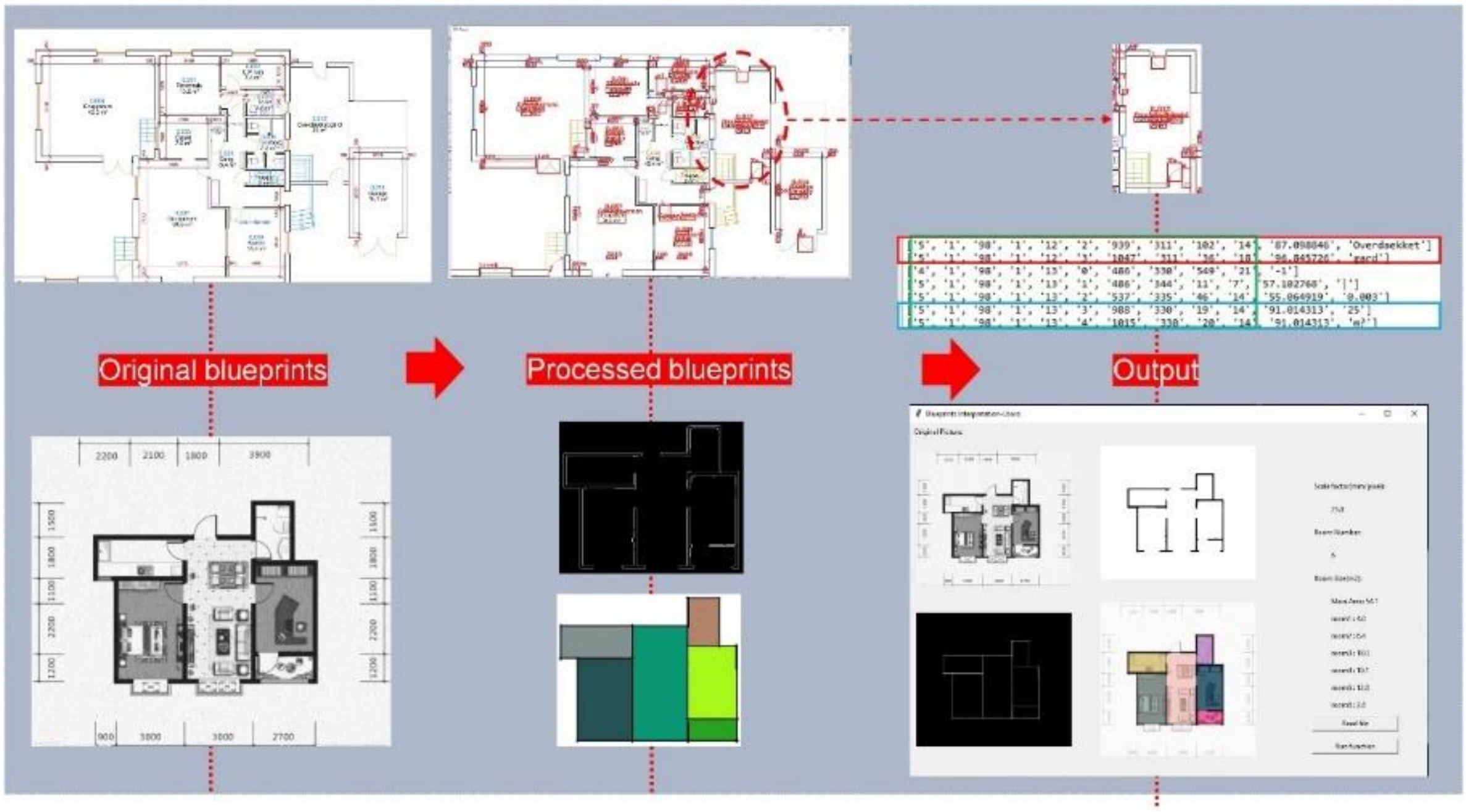

Fig. 4. Example of blueprint interpretation

The prospect that emerges from this work is by no means the replacement of skilled experts in insurance risk assessment. The reality is far from the media Frankenstein myth about AI growing to become more intelligent than its human creators. The prospect is the development of tools support to risk engineers and enhance their work.

Over time these tools could become very powerful. Imagine a risk engineer, able to prepare for a visit more rapidly with an AI-prepared summary of features from a blueprint. Equipped with a camera, standard features might be captured automatically, freeing their attention to the customisation of their risk assessment, focusing on the unusual, the differences from what might be expected that could be crucial in terms of a fire spreading.

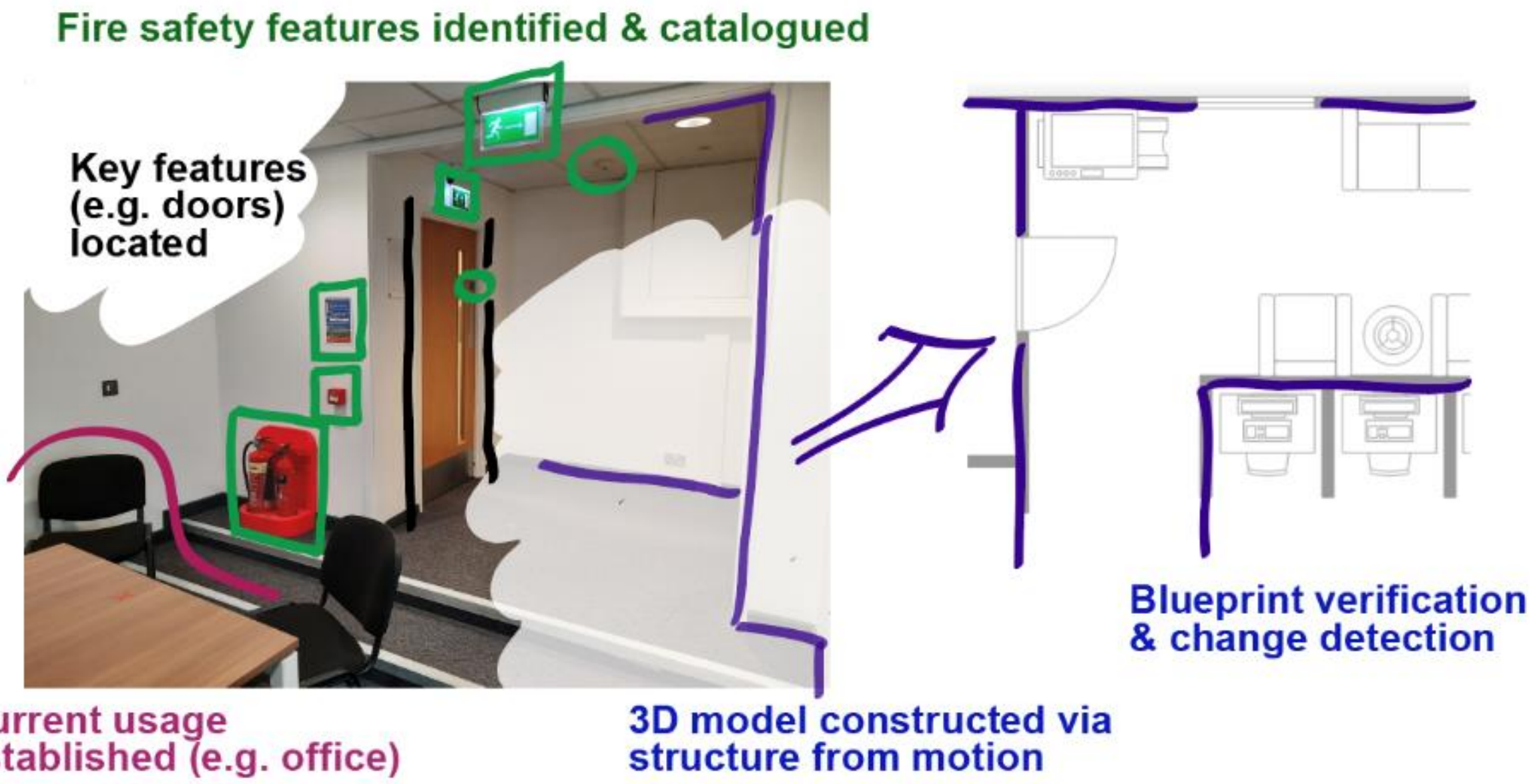

Figure 5 - Figure of an idealized future outcome.
A future tool might identify key features (e.g., door, fire extinguishers), and self-generate a 3D model and thus a floor plan to verify blueprints and highlight changes since construction.

## 4. Conclusions: future research

This paper reports the outcome of an initial pilot study of the use of artificial intelligence software (AI) to extract information from building blueprints, information that could then be used in risk assessment for insurance of industrial and commercial property. It uses public domain residential property blueprints and focuses on calculation of room area as a sample output. The technical contribution is demonstrating how a combination of machine learning tools can be incorporated into an integrated end to end process: beginning with the pre-processing of blueprints; distinguishing and locating boundaries such as walls and other features and objects recorded on the blueprint; identifying these features and objects by matching them to stored templates; extracting textual information and identifying what it represents; and finally using the extracted information to produce an analytical output.

There is no inherent barrier to extending these same methods to blueprints for industrial and commercial properties, with sufficient access to blueprints and time spent on developing and training the software. Such a tool can be capable of rapidly extracting a variety of blueprint information, relieving risk engineers of much routine work. This can allow them to focus on where they add most value, using their professional knowledge and experience to review the outputs of automated blued print interpretation and then combining these outputs together with other information to prepare property insurance risk assessments.

At Loughborough we have conducted extensive work on technology adoption in insurance, over the course of the past three years, in our TECHNGI research project. We conclude this paper with a brief discussion, drawing on our TECHNGI work, of some of the main issues that will need to be addressed for successful adoption of automated AI-based information extraction and other digital tools in industrial and commercial property insurance. Perhaps the most important lesson here is that the software itself, such as the machine learning software used in the proof of concept work reported here, is only one element in the broad task of using digital tools for automated information processing and business analysis in industrial and commercial property insurance.

- Software application has to be tailored to business context. In the case of blueprint information extraction for industrial and commercial properties, there will need to be a quite extensive experimentation to establish the feasibility of commercial application, examining both the very wide range of properties for which insurance policies are provided and the substantial variation in the format of property blueprints.
- There will inevitably be limitations on what can be extracted from blueprints, this is especially likely for older properties. Further research is needed to assess the possibilities of supplementing blueprints with additional digitally processed data, most obviously using tool of computer vision applied to videos and still photographs, to enhance blueprint processing.
- Alongside these technical investigations, equally important will be extensive engagement with risk engineers, to adequately understand their use of information and where automated data processing can most usefully enhance their work. While current business practice in industrial and commercial property insurance is still largely manual, this does not mean that risk assessment is ineffective. The case needs to be made to practitioners of the benefits of automated data processing. For this fuller investigation is needed into (a) opportunities for savings of cost and time i.e. maintaining the existing practice but assessing how much time and can be save through automated data processing; (b) opportunities for improving existing practice, e.g. using automated tools to simulate variation in hazard and consequences for financial loss; (c) exploring the potential for greater standardisation of metrics of property insurance risks, based on initial standardised calculations based on automated processing of blueprint and other information which can then be adjusted based on the professional assessments of experienced risk engineers; (d) incorporating fully

- A more systematic 'ontology based' approach, i.e., one with industry level agreement on definitions of risk data and on the computation for standardised risk metrics, can support broader automation of the provision of quotations for insurance policies and support a shift from standarised to more tailored cover. This can also support 'knowledge management' i.e., the recording and dissemination of professional skills so that the understanding of risk engineers is more widely shared with their professional colleagues and embedded in professional training (not left entirely to 'on the job' apprentice style individual to individual acquisition).
- The managerial, strategic and economic implications of adoption of digital technologies in commercial and industrial property insurance also need examination. The development of more standardised metrics of risk may benefit customers by supporting greater competition in underwriting and hence reducing premiums. Standardisation may also facilitate greater comparability of risk and hence support greater sensitivity of premiums to investment in loss mitigation.

Fully achieving all these possible benefits, which go well beyond operational cost savings, is likely to require a coordinated cross industry adoption. The benefits are unlikely to be fully realised if adoption of digital technologies for automated property insurance risk assessment is left as the un-coordinated choice of individual risk engineers or the departments in which they work. The need for co-ordinated adoption further suggest that there will considerable value for insurance clients from comprehensive investigation of the full range of benefits from adoption of digital tools for risk assessment in industrial and commercial property insurance.

On vulnerability $V$ - identifying for example proximity of different structures that can allow the spread of fire, or the presence of fire walls and fire doors, are an essential part of the estimation of exposure to fire loss. The further identification of active prevention such as sprinkler systems can support estimation of vulnerability.

This automation will though not be a replacement for the judgements of an experienced risk professional – they will be a supporting tool that provide an initial baseline of information and routine calculations that can save on the amount of manual information processing that is required to for assessing exposure $E$ and vulnerability $V$.

# Appendix A -- Literature review on risk assessment

Property risk can be defined as the risk of financial loss arising from owning a property. Fuller definition is provided by (Advisen Ltd., 2013): "The term 'property risk' refers to risk events that specifically impact an organization's facilities and other physical infrastructure. Risk events such as fires, adverse weather conditions, and terrorist attacks all fall into the category of property risk. In addition to damaging and destroying physical property, property risk events also have the potential to create stoppages in business operations and material financial losses."

Focusing on physical damage, it is usual to distinguish four categories of risk:

1. Natural catastrophes
2. Fire and explosions
3. Failure or malfunction of building installations
4. Criminal damage

For all these categories of risk, a standard approach is to quantify risk $Q$ as the probability of the hazard event $P$ multiplied by exposure $E$ and by vulnerability $V$.

$$Q = P \times E \times V$$

Here:

- The probability parameter ($P$) is the probability of the given event occurring e.g., 1 in 50 years.
- The exposure parameter ($E$) is the potential economic cost of the damage if the event occurs. In the case of a fire this could be
- The vulnerability parameter (V) measures the extent to which the event results in financial loss, allowing for loss prevention technology, mitigation and other factors that reduces or increase the financial costs of the event.

Whilst useful, this framework is a very significant simplification. A fuller description of risk will recognise a distribution of both hazard (i.e., varying severity, frequency, spatial pattern) and of loss. This short coming is most obvious in the case of natural hazards such as floods or storms, since these can very considerably in severity, from mild to extreme, and with the consequence that both exposure and vulnerability and hence also financial loss vary with the severity of the event. For example a building may be on sufficiently high ground that it is not exposed at all to a flood of 2 metres in depth above some baseline, that might occur once in 50 years, but is substantially exposed to a flood of 5 metres in depth that might occur once in 500 years.

Insurance of the risk of natural hazards must also take account of correlation of hazard events between different properties. If many insured properties are in the same or proximate locations, with a common exposure to hazards such as a major flood or storm, then this can limited diversification of risk in an insured portfolio, perhaps necessitating the purpose of some reinsurance. For other risks though, such as those associated with fire and failure of building installations, correlation of hazard is of little concern and reinsurance should not be needed.

In what follows we focus on risk of fire and explosions, drawing on the literature review in (Sølvsten, 2022).

Dealing first with exposure $E$. (Holborn, Nolan, & Golt, 2004) highlight the wide range of severity of potential fire events. Although, relatively few lead to substantial damage, larger events can account for half or more of all losses. It is therefore sensible to at least distinguish the possibility of an extreme fire event of low probability $P$ with widespread damage (i.e., high exposure $E$) and a more moderate event with a greater probability $P$ but lower exposure $E$.

Turning to vulnerability $V$, this depends on fire loss-prevention technology, which can be divided into two main groups:

1. *Passive fire protection*: This is defined as materials or structures such as fire walls which provide fire protection without the need to be activated or switched on. They compartmentalise the building into zones in order to limit the spread of fire.
2. *Active fire protection*: This comprises system such as sprinkler systems which only gives protection against fire once activated. It can be activated either by detection technology or manually. Generally used for the purpose of extinguishing fires.

These should be used together to provide fuller protection against damages and financial loss, however, interest in active techniques has been increasing due to its ability to be customised.

These active fire protection techniques work to limit the spread of fire and therefore reduce the vulnerability. They can be further divided down into:

1. *Fire suppression systems*: Any system that works to control and contain a fire until the fire is extinguished or for rescue to arrive. This includes automatic closing fire doors or most commonly, sprinkler systems (a commonly used type of fire suppression system that are usually automatically triggered when sensing a temperature of around 60°C-80°C (Figueroa, 2013).
2. *Fire suppression systems*: Automatically detecting a fire and trigger a response or action. For example, a smoke detector causing an alarm to sound or fire doors to automatically shut.

There is research on the effectiveness sprinkler systems, but not on other forms of active fire protections. This research has used one of two approaches. Either a component-based approach, testing the functionality for each component and then seeks to model their combined effectiveness in the event of a fire; or a system based approach that uses data from previous fire events to evaluate the effectiveness of the whole system. (Frank, Gravestock, Spearpoint, & Fleischmann, 2013) reviewed 15 studies which used the system-based approach and conclude that the effectiveness of a sprinkler system in limiting damage to be between 70.1% and 98.8%,

Automated extraction of information from building blueprints could potentially be used for assessing both exposure $E$ and vulnerability $V$ to fire loss. The extraction of information, described in Appendix C, identifying for example proximity of different structures that can allow the spread of fire, or the presence of fire walls and fire doors, are an essential part of the estimation of exposure to fire loss. The further identification of active prevention such as sprinkler systems can support estimation of vulnerability.

This automation will though not be a replacement for the judgements of an experienced risk professional – they will be a supporting tool that provide an initial baseline of information and routine calculations that can save on the amount of manual information processing that is required to for assessing exposure $E$ and vulnerability $V$.

# Appendix B – Existing literature on applying AI to building blueprints

## Applying AI to building blueprints

The automated and intelligent interpretation of building blueprints is an emerging research field, which is related to massive applications in the architectural domain. In terms of the interpretation and analysis of building blueprints, numerous rule-based heuristics methods and image processing methods were previously applied to the detection and classification of structural primitives in the blueprints. Nowadays, machine learning and artificial intelligence approaches (e.g., fully connected neural networks and CNN) are widely involved in the vectorization, segmentation and detection of building blueprints (Paudel, Dhakal, & Bhattarai, 2021).

Previously, the low-level image processing was used in some traditional methods for the element recognition. For example, Ryall et al. firstly applied a semi-automatic method for the room segmentation in 1995 (Ryall, Shieber, Marks, & Mazer, 1995). Other early methods locate walls, doors, and rooms by detecting graphical shapes in the layout, e.g., line, arc, and small loop. Dosch et al. introduced a complete platform for analyzing architectural diagrams. Herein, numerous automated graphics recognition processes are used for the basic primitives recognition (Dosch, Tombre, Ah-Soon, & Masini, 2000). Siu-hang et al. transformed bitmapped floor plans to vector graphics and generated 3D room models (Siu-hang, Wong, Yu, & Chang, 2009). Ahmed et al. made efforts to separate text out from graphics and extract lines with various thickness, where primitives are extracted from thinner lines and walls are supposed to have thicker lines; Furthermore, these information are used to locate doors and windows (Ahmed, Liwicki, Weber, & Dengel, 2011). Gimenez et al. delivered a comprehensive review about recognizing geometry, topology and semantics information of building elements from 2D scanned plans, and generating 3D building information models (Gimenez, Hippolyte, Robert, Suard, & Zreik, 2015). Other interesting works include the one by Llados et al., who uses the Hough Transform to identify primitives from hand-made architectural blueprints (Lladós, López-Krahe, & Martí, 1997), and the one by Lu et al., who tackle the structural-objects recognition and wall-shapes classification (Lu, Yang, Yang, & Cai, 2007). The Hough Transform is the fundamental approach for recognizing geometric entities and objects from blueprints. Moreover, over 2500 papers have been published to improve it (Zhao, Deng, & Lai, 2020). Nowadays, massive developed algorithms based on the popular Hough Transform have been proposed, such as Random Hough Transform, Digital Hough Transform, Probabilistic Hough Transform, Progressive Hough Transform, Fuzzy Hough Transform and Generalized Hough Transform.

Recently, some AI approaches have been widely applied to blueprints interpretation. Liu et al. modified and trained a DNN to recognize junction points in a blueprint, and then utilize integer programming to join the junctions to locate the walls in the blueprint (Liu, Wu, Kohli, & Furukawa, 2017). The domain of image style transfer has enhanced by the growth of generative network models. Huang and Zheng introduced a way of implementing GAN in the field for the recognition and generation of floor plans. They parsed floor plans by segmenting areas with different functions to design by data (Huang & Zheng, 2018). S. Kim et al. proposed a method to convert diverse floor plans into an integrated format also based on GAN in the vectorization process (Kim & Park, 2018). Sharma et al. applied the GAN convolutional network for blueprint detection and retrieval (Sharma, Gupta, Chattopadhyay, & Mehta, 2019). The CNN model was used by Jiang et al. to recognize pixels on the wall and the door (Jang, Yu, & Yang, 2020). Also, they used the extracted semantic information to reconstruct the indoor plan from the floor plan. It is well-known that AI models commonly need large amounts of data to train the model, and bigger datasets can normally yield better results. The datasets are also the crucial prerequisites for the AI-enabled blueprints interpretation. However, the dataset preparation and processing are quite time consuming. Kalervo et al. made a new dataset (5000 images) of blueprints for buildings' interior modelling, and they employed CNN to detect and analyse blueprints (Kalervo, Ylioinas, Häikiö, Karhu, & Kannala, 2019). And several popular works regarding the usage of AI on blueprints interpretation would be specifically indicated below.

(Yamada, Wang, & Yamasaki, 2021) develops a graph structure extraction approach from blueprints. This approach first recognises the contents on blueprints by pixel-wise semantic segmentation using deep learning (DL). Then, a rule-based transformation of the images is performed with semantic labels into graph structures. Herein, rooms are represented as nodes in the graph model, and the linkage of these rooms are further represented as edges. The generated graph can reflect the original blueprints with essential semantic information. The detailed two-step procedure is displayed in Fig.B1.

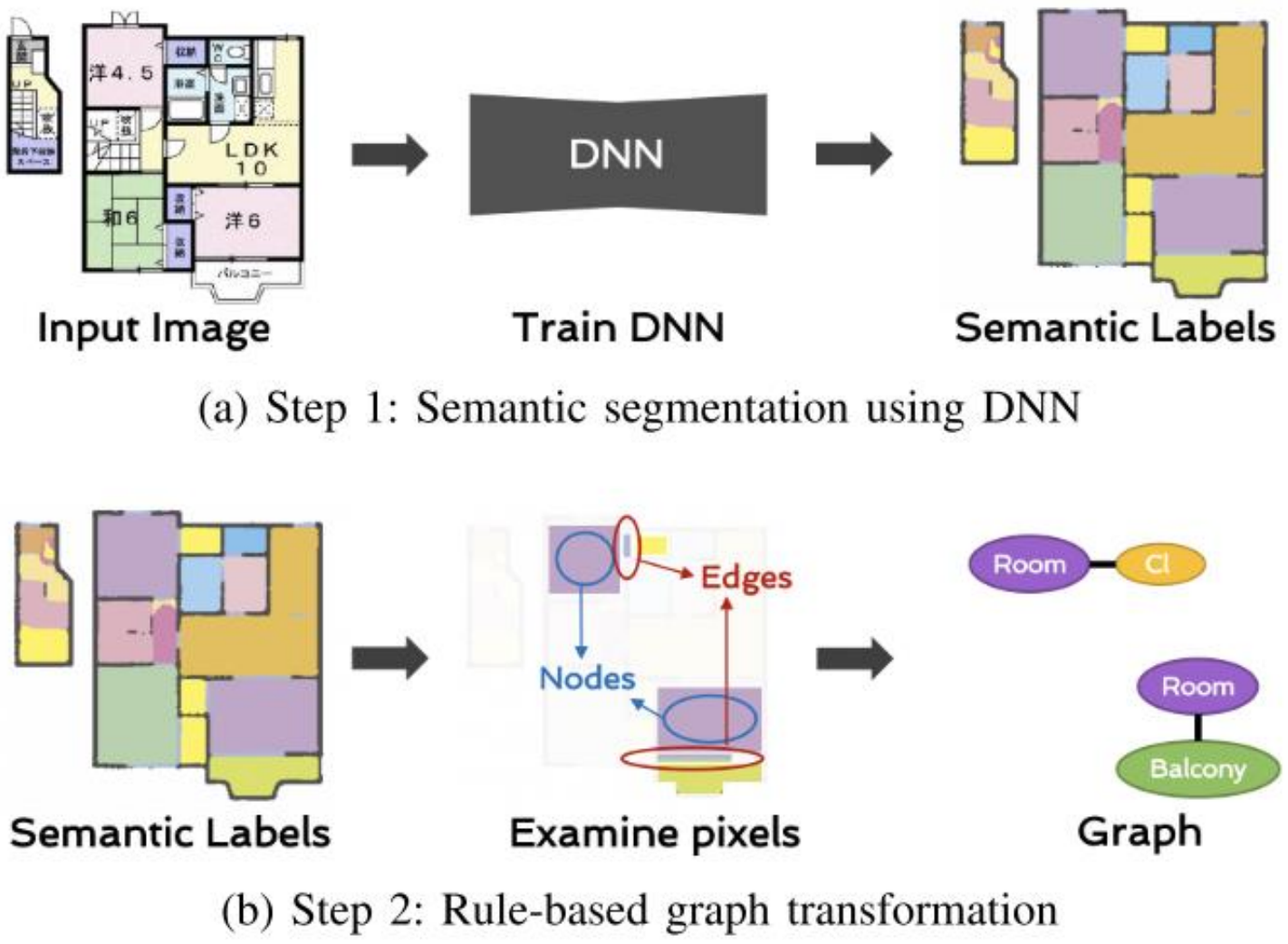

Fig. B1. The graph structure extraction method

Raster-to-Vector (R2V) originally produced a transformation approach for converting raster blueprints to vector-graphics representations, as shown in Fig. B2. The interpretation of blueprints by R2V (Liu et al., 2017). The proposed approach invented two intermediate representations integrating both geometric and semantic information of a blueprint rather than relied on low-level image processing heuristics.

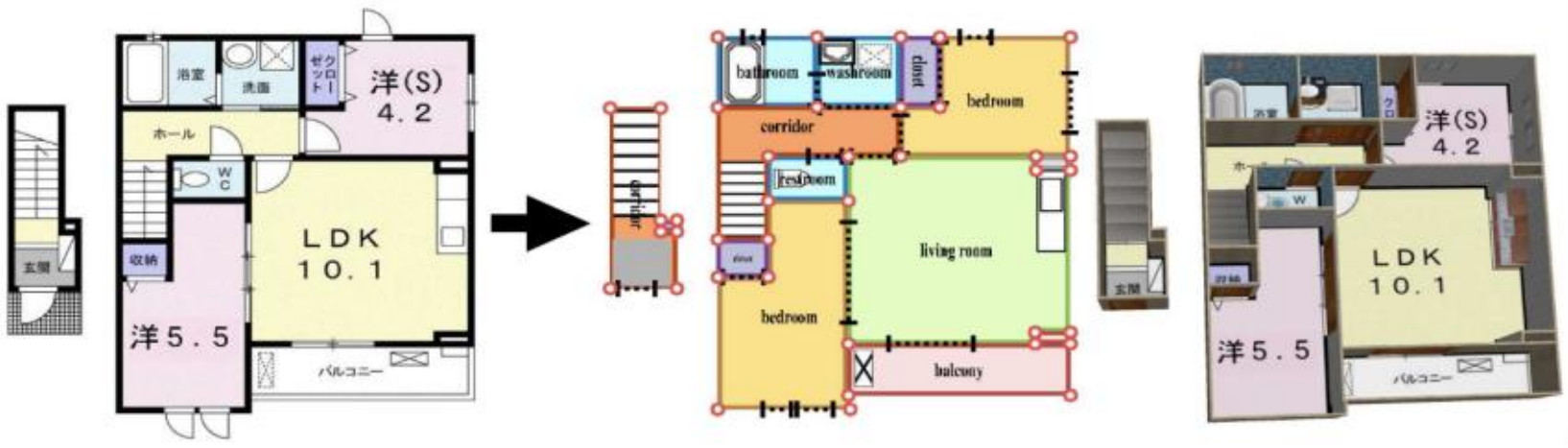

Fig. B2. The interpretation of blueprints by R2V

Specifically, the detailed transformation process is described in Fig. B3. The detailed transformation processes of R2V. A neural architecture (i.e., CNN) first transforms a rasterised image to a set of junctions that represent low-level geometric and semantic information (e.g., wall corners or door endpoints). Then, via integer programming, junctions are aggregated into a set of simple primitives (e.g., wall lines, door lines, or icon boxes) to produce vectorised blueprints with high-level geometric and semantic constraints. Finally, post-processing is utilised to convert the blueprint data into vector-graphics representation. The residual part of the detection network is borrowed from (Bulat & Tzimiropoulos, 2016). The used dataset in this project is ResNet-152 (He, Zhang, Ren, & Sun, 2016). Some interesting and popular works have been summarized below in Tab. B1.

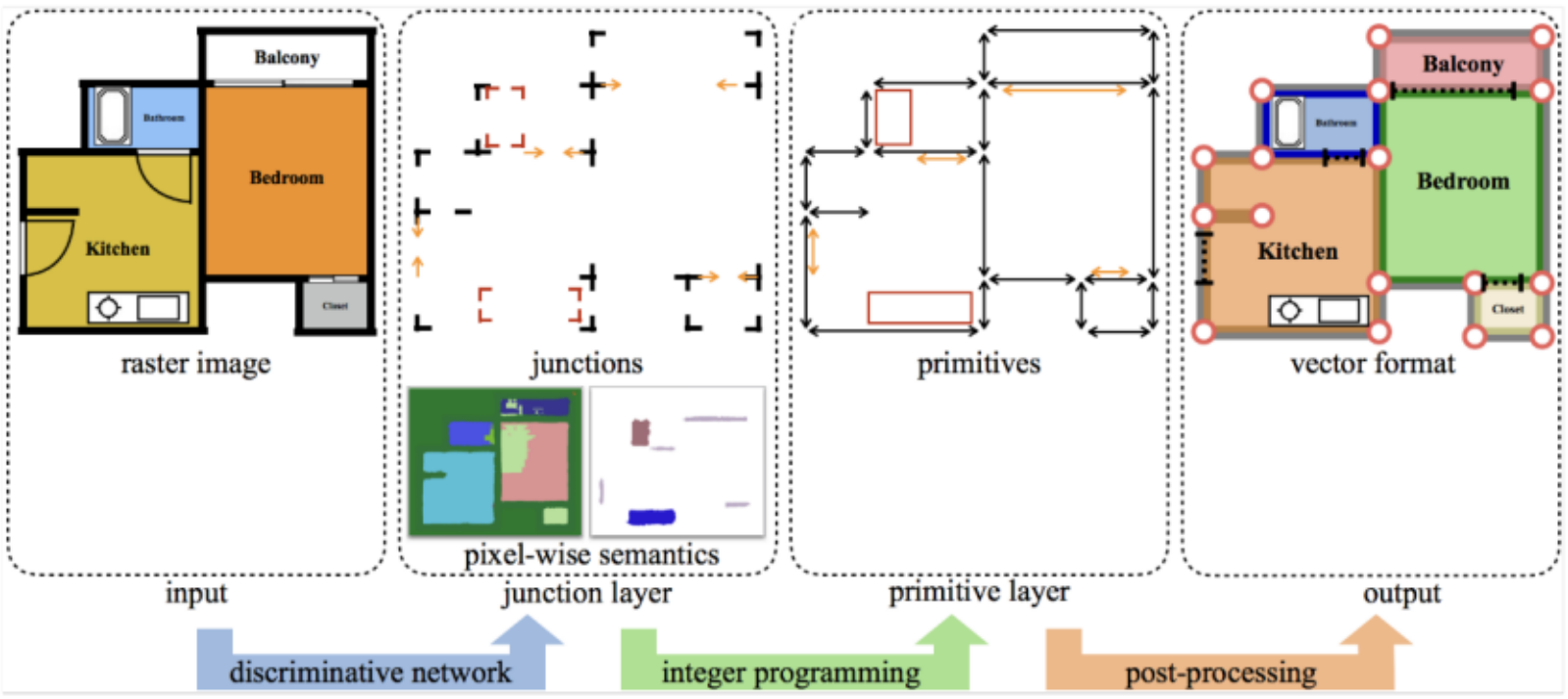

Fig. B3. The detailed transformation processes of R2V

| AI projects for blueprint interpretation | Year | Information Capturing | Main algorithms | Generated primitives | Final output | Dataset |
|---|---|---|---|---|---|---|
| Raster-to-vector (R2V) | 2017 | Geometric information, Semantic information | CNN, Integer programming | Wall lines, Door lines, icon boxes | Vector-graphics representation | LIFULL HOME'S (4800) |
| Deep floor plan recognition | 2019 | Room type, Room boundary | Deep multi-task neural network, VGG | Walls, doors, windows, room regions | Room types | R2V (815 images) and R3D (214 images) |
| Automatic recognition and segmentation of architectural elements | 2020 | Architectural elements | FCN | Door, window, wall, column, stair, background with different colours | Predicted output image | Manual labelling on 300 images |
| Graph Structure Extraction | 2021 | Semantic labels | DNN model DeepLabv3+, rule-based graph extraction | Wall, dinning room, kitchen, bath room, entrance, stairs, doors, etc | Transformed graph with nodes (rooms) and edges (adjacence) | LIFULL HOME'S (4800) + manual labelling by LabelMe |
| Room classification | 2021 | Room area, length, width, number, parent or child room | MLP, GNN | Living room, kitchen, bedroom, corridor, closet, etc | Undirected graph | House-GAN (143184) |

Tab. B1. The summary of some AI-driven approaches

## Further discussion of AI methods

Nowadays, AI has made its place in the blueprint interpretation. Numerous AI approaches (e.g., Support Vector Machine (SVM), Convolution Neural Network (CNN), Deep Neural Network (DNN), Generative Adversarial Network (GAN), etc.) have been specially used to recognize the blueprints. Herein, deep learning (DL) is one of the research hotspots in the scope of blueprints interpretation. In terms of the detection process, DL algorithms can be classified into two types: two-stage objection detection and one-stage object detection.

The basic architecture of two-stage detector includes regional proposal network to feed region proposals into classifier and regressor, such as Region-based CNN (R-CNN), FAST R-CNN, FASTER R-CNN, MASK R-CNN, etc. The basic architecture of one-stage detector can predict the bounding boxes from input images directly, such as You Only Look Once (YOLO), YOLOv2, YOLOv3, YOLOv4, YOLOv5, SSD, DSSD, RetinaNet, M2Det, RefineDet, etc. The development of related CNN algorithms is summarized below in Fig. B4.

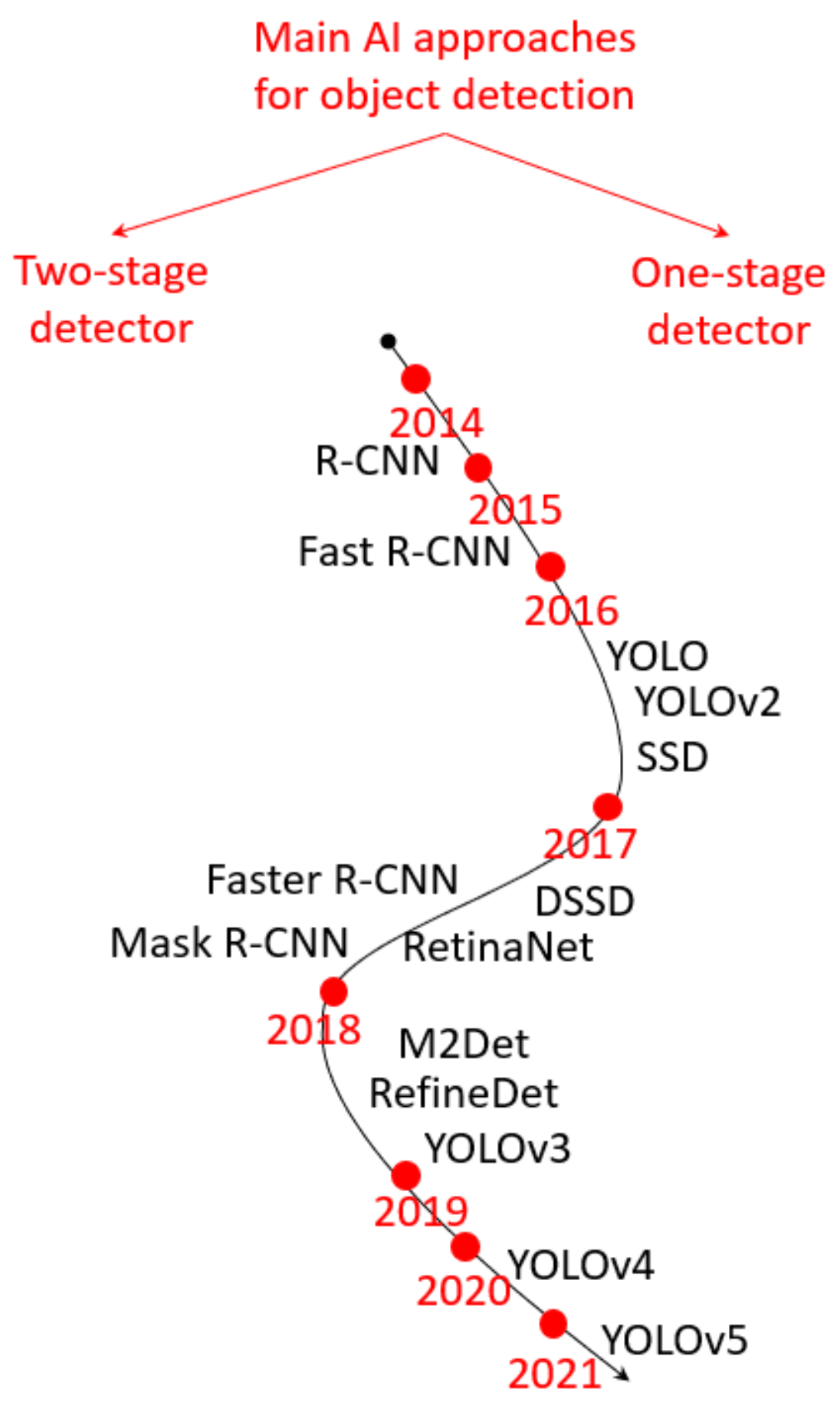

Fig. B4. The brief summary of CNN

Note: One stage methods are shown to the right of the time line, two stage convolution methods are shows to the left of the time line.

R-CNN or RCNN, is an extension of CNN, and a classic type of ML model for CV tasks, especially for object detection. R-CNN detector consists of four modules. The first module generates category-independent region proposals. The second module extracts a fixed-length feature vector from each region proposal. The third module is a set of class-specific linear SVMs to classify the objects in one image. The last module is a bounding-box regressor for precisely bounding- box prediction. Because R-CNN conduct a ConvNet forward pass for each region proposal without sharing computation, R-CNN would take a long time on SVMs classification.

Fast R-CNN drastically improves the training and detection time from R-CNN. It extracts features from an entire input image and then passes the region of interest (RoI) pooling layer to get the fixed size features as the input of the following classification and bounding box regression fully connected layers. The features are extracted from the entire image once and are sent to CNN for classification and localization at a time. Compared to R-CNN which inputs each region proposals to CNN, a large amount of time can be saved for CNN processing and large disk storage to store a great deal of features can be saved either in Fast R-CNN. As mentioned above, training R-CNN is a multi- stage process which covers pre-training stage, fine-tuning stage, SVMs classification stage and bounding box regression stage.

Fast R-CNN uses selective search to propose RoI, which is slow and needs the same running time as the detection network. Faster R-CNN replaces it with a novel RPN (region proposal network) that is a fully convolutional network to efficiently predict region proposals with a wide range of scales and aspect ratios. RPN accelerates the generating speed of region proposals because it shares fully-image convolutional features and a common set of convolutional layers with the detection network. Furthermore, a novel method for different sized object detection is that multi-scale anchors are used as reference. The anchors can greatly simplify the process of generating various sized region proposals with no need of multiple scales of input images or features.

Faster R-CNN is deep convolutional network for object detection which show a single, end-to-end and unified network to users. Mask R-CNN is an extension to Faster R-CNN for instance segmentation task. Regardless of the adding parallel mask branch, Mask R-CNN can be seen a more accurate object detector.

# Appendix C – our proof-of-concept work on AI applied to blueprints

This Appendix summarises the workflow that emerged from our initial proof-of-concept modelling for automated blueprint analysis. In broad terms this consists of the following main steps. First preprocessing, which is itself has two main stages (i) simplification make walls into lines. (ii) clarification, to remove noise, grey (overview). Then extracting shapes and text. Finally output of identified shapes with associated names.

This Appendix is organised as follows. C.1 is an overview of the methods involved. C.2 is the blueprint preparation for the interpretation. C.3 delivers two ways for the template matching. C.4 introduces the OCR technology into the blueprint interpretation. C.5 bundles the aforementioned processes and introduces a unified platform for the blueprint interpretation.

## C1. Overview of the proposed solution

Fig. C1 describes our proposed solution with top-down steps. The proposed solution starts from the blueprints library preparation. From the blueprints library, the pending blueprint is inputted as the image in our solution. Then, the blueprint interpretation is mainly triggered by the pre-processing stage in our solution. Following the pre-processing, our solution will act on the contour processing, room processing and object detection. Herein, the contour processing and room-related processing are the key parts in our solution. The contour processing is corresponded to the macro blueprint interpretation, and the room-related processing is applied to the micro blueprint interpretation. The following sections will particularly introduce and demonstrate these steps.

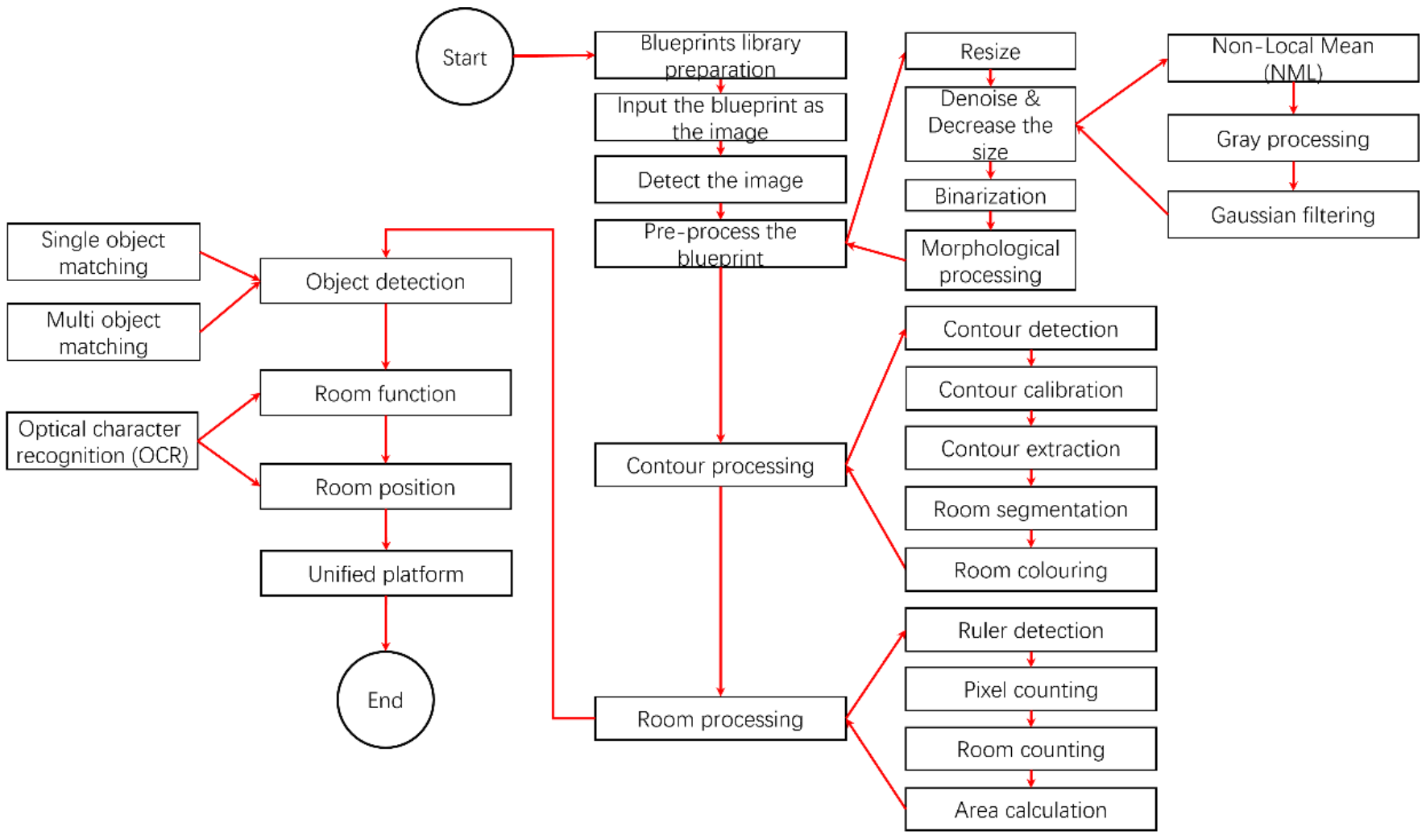

Fig. C1. The overview of the proposed method.

## C2. Blueprint pre-processing

Considering the diverse qualities and standards of blueprints, the preprocessing of blueprints is of importance for blueprint interpretation. As shown in Fig. C2, the original blueprint is preprocessed in a systematic order, which consists of resizing, denoising, size reduction, binarization and morphological processing. By pre-processing the blueprints, the quality of image is effectively improved. Specially, the noise of the blueprint can be reduced, and the features which are necessary for the blueprint interpretation can be enhanced.

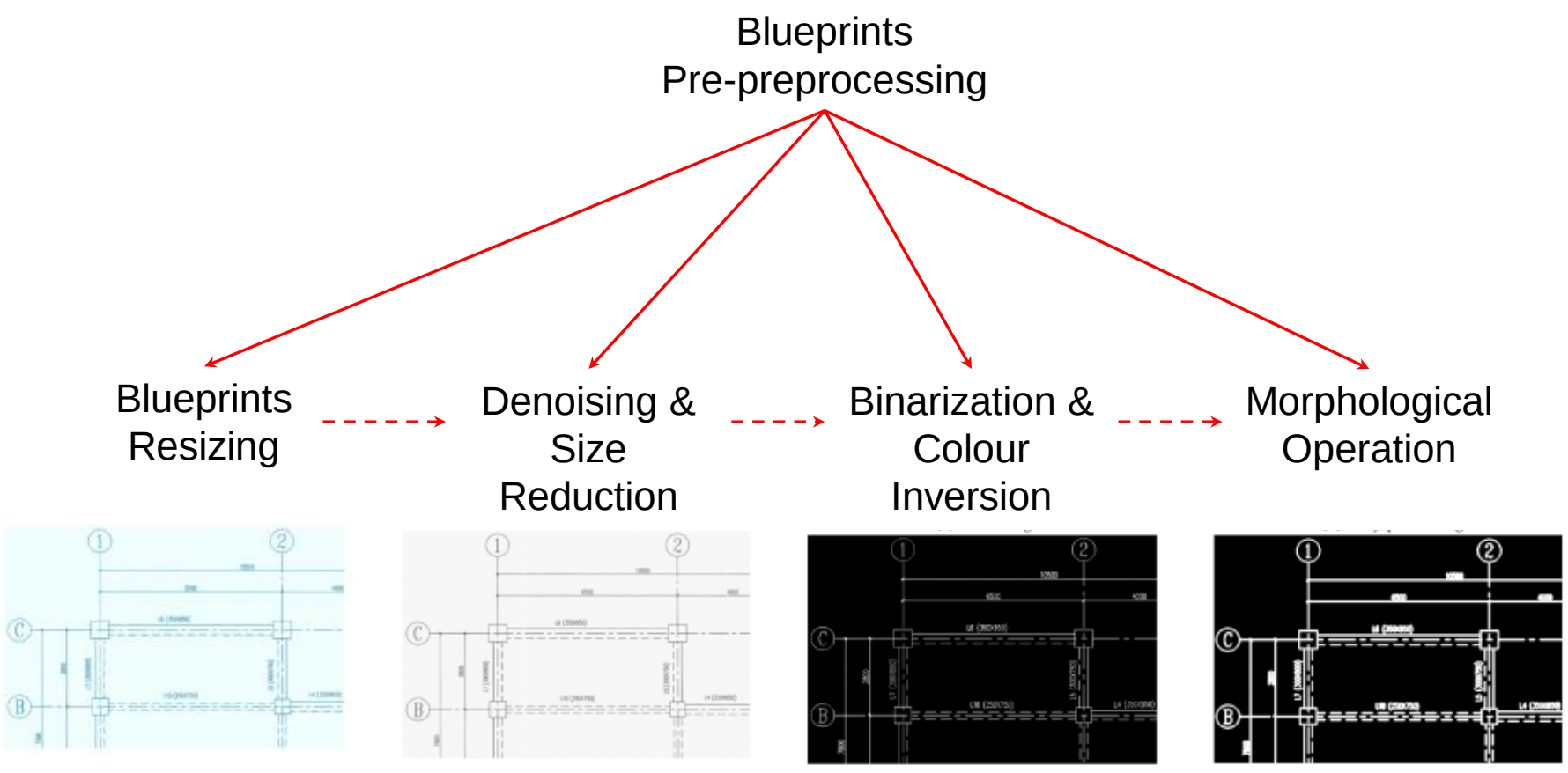

Fig. C2. The pre-processing stage of blueprints interpretation

These four elements of preprocessing are now each described in more detail.

a) Blueprints Resizing

By resizing the image of a blueprint, the physical size of an image can be changed without altering the pixel dimensions in the image. Only the width and height of the image are changed. Based on the requirement, the original and latest aspect ratios can be preserved for the following processing. The blueprint resizing would be useful for the batch processing.

b) Denoising and Size Reduction

The NML and Gaussian filtering are utilized in our solution for image denoising. Different from “local mean (LM)” filter, NML takes a mean of all pixels in the image, weighted by how similar these pixels are to the target pixel. This results in much greater post-filtering clarity, and less loss of detail in the image compared with local mean algorithms. Furthermore, Gaussian filtering is applied to the image denoising. This type of filtering is easy to implement with automatic censoring. It is the most popular filtering way in computer vision algorithms for the purpose of enhancing images structures in various scales. As shown in Fig. C3, the original blueprint is blurred by Gaussian filtering.

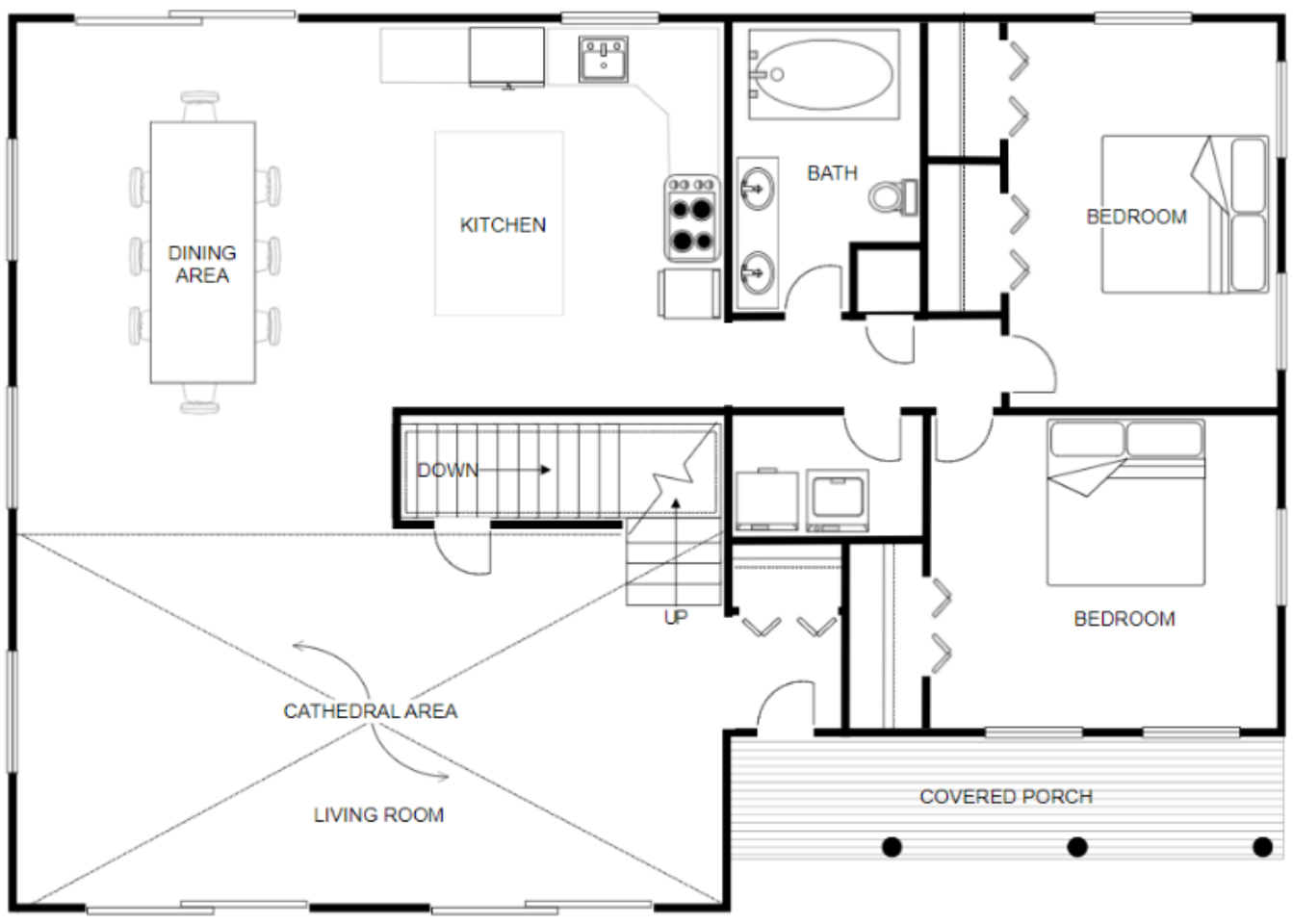

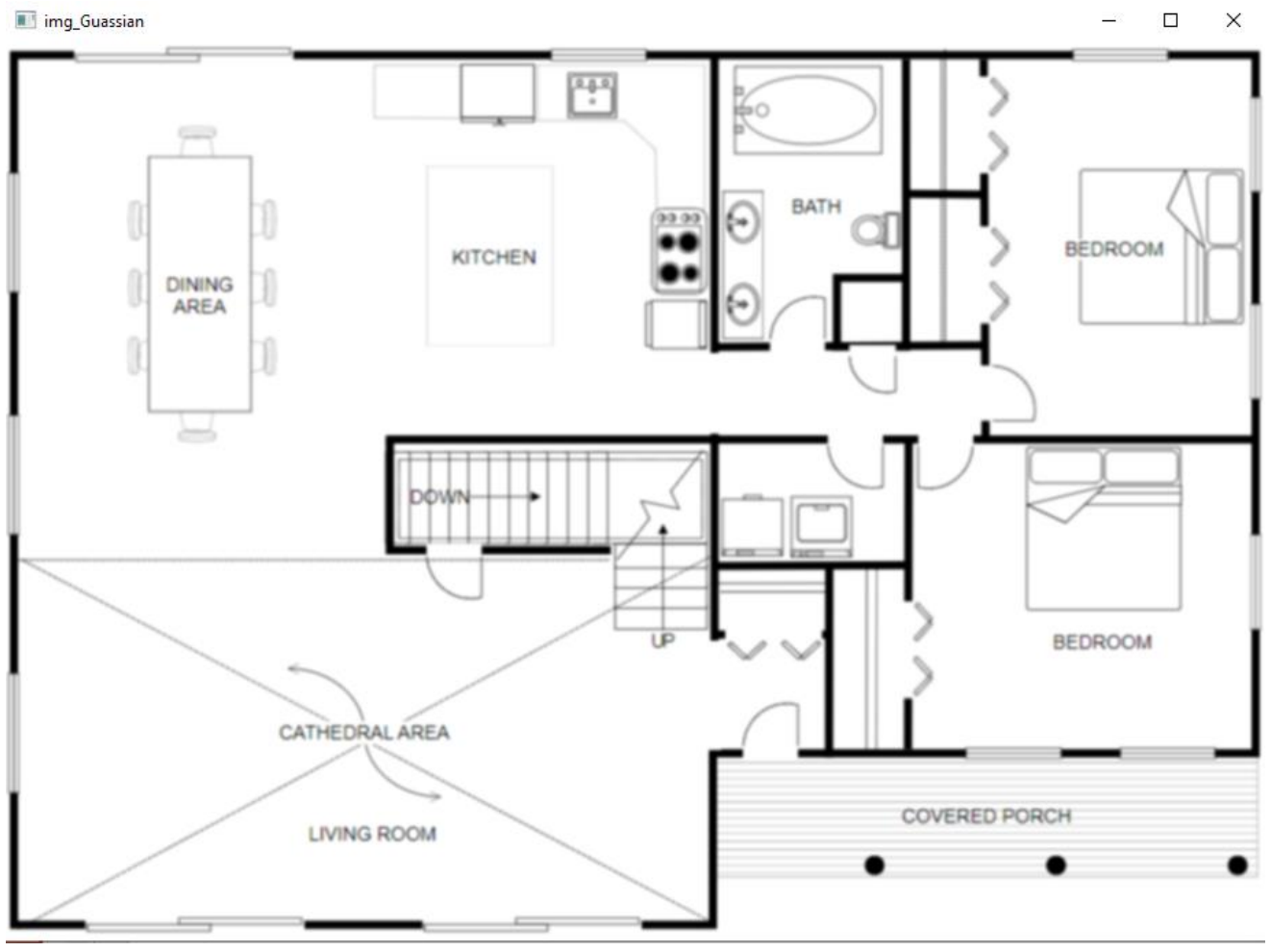

Fig. C3. The Gaussian filtering of blueprints.

Gaussian filtering, which is also known as Gaussian smoothing, uses the Gaussian function to blur the image. The noise of an image can be typically reduced by the Gaussian filtering. Based on the various requirements, the degree of smoothing can be determined by the standard deviation of the Gaussian.

Compared with other filtering methods, such as Mean filtering, Median filtering and bilateral filtering, Gaussian filtering is much better at separating frequencies by using the sigma and window size. This type of filtering technology is widely utilized as a pre-processing stage in computer vision algorithms for the purpose of enhancing images structures in various scales. For example, Gaussian filtering is commonly applied to the edge detection. Because most edge-detection algorithms are sensitive to noise, the noise reduction by Gaussian filtering can effectively improve the edge-detection results.

Then, the gray processing is used to reduce the size of an image. As shown in Fig. C4, grey processing can convert the RGB image to the grayscale image, which can greatly reduce the image size without the distortion. The grayscale image contains only white, black and grey colours with several levels.

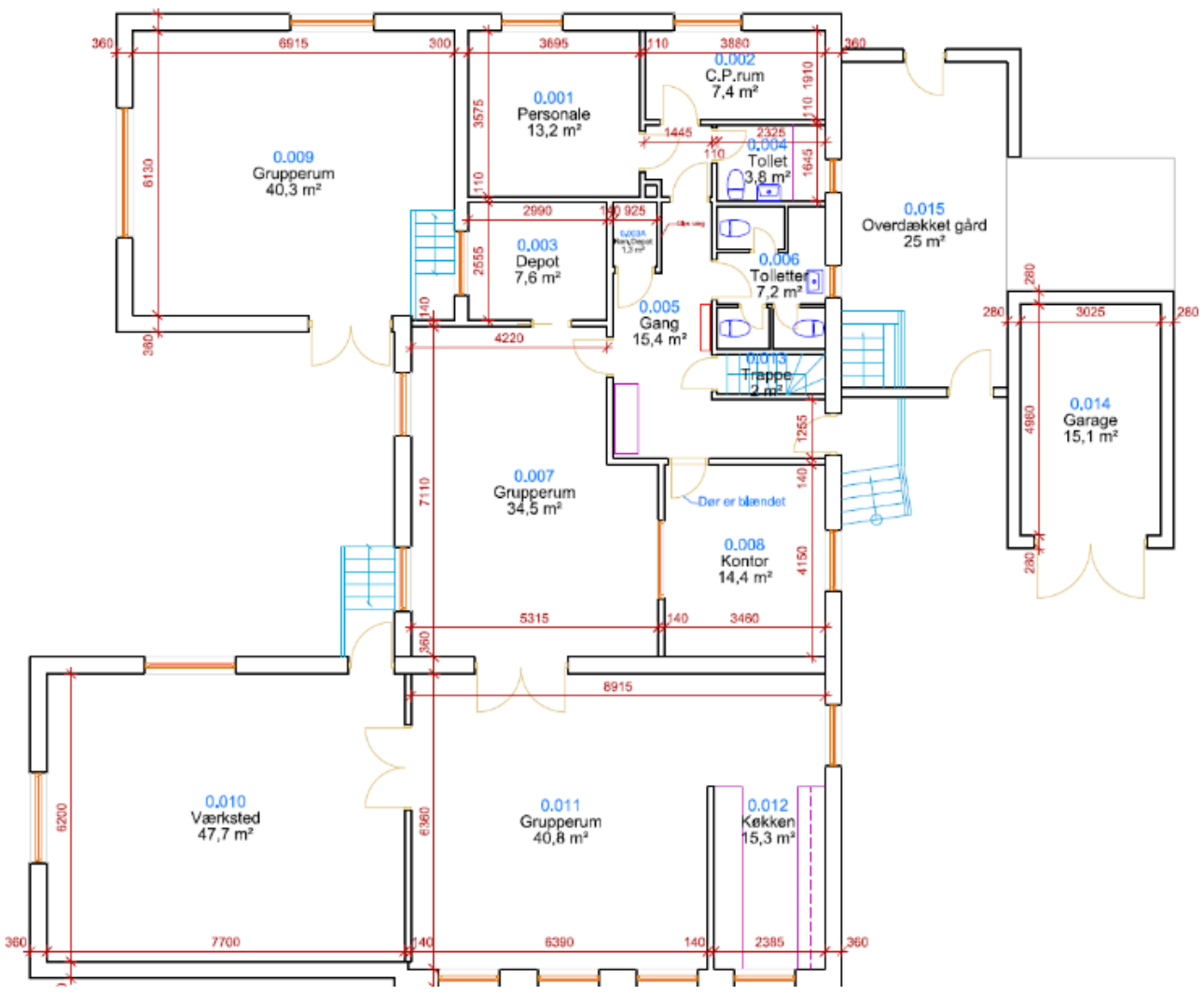

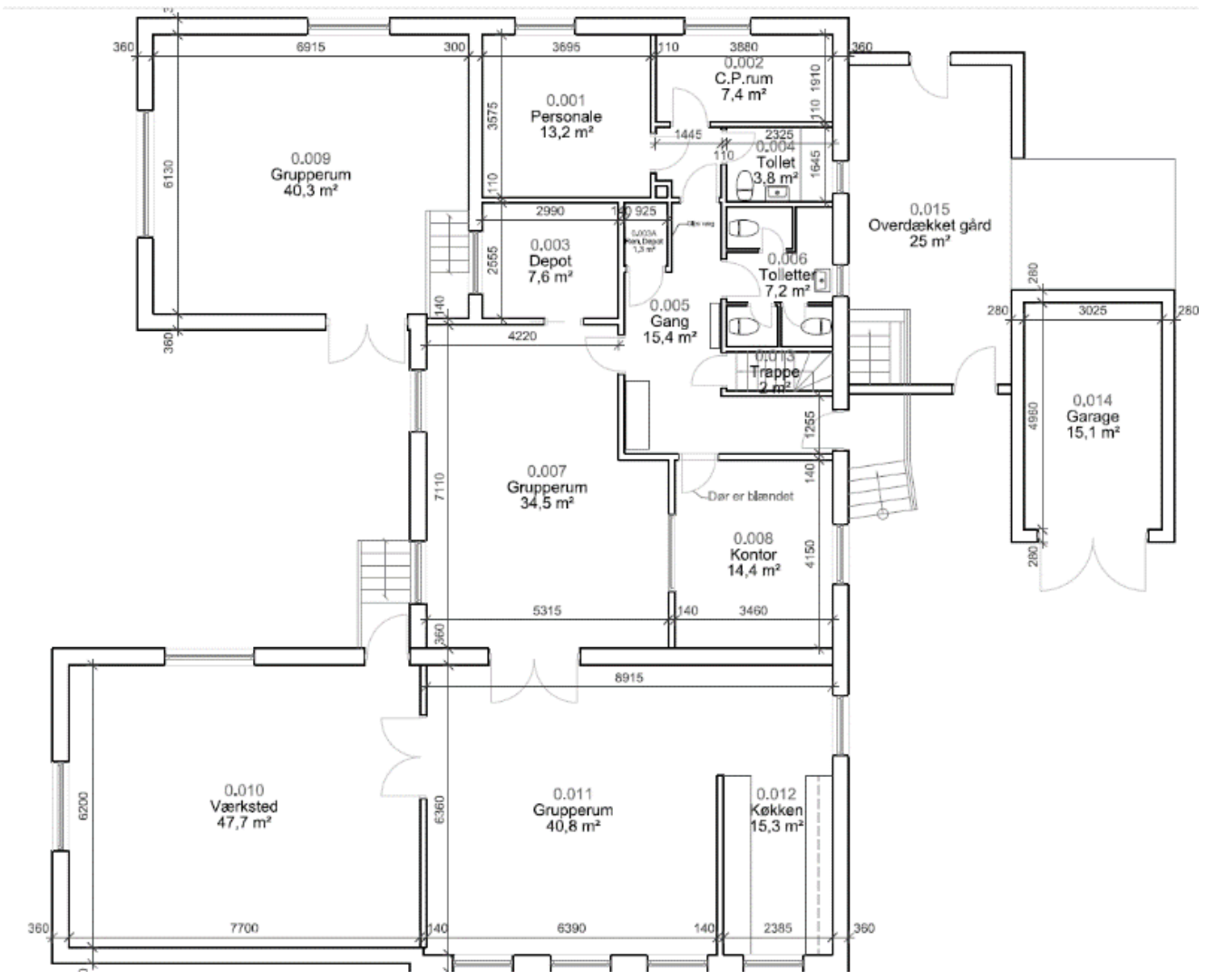

Fig. C4. The gray processing of blueprints

The grey processing of colour images is commonly utilized for descriptor extraction because the grayscale image can greatly reduce computational cost and simplify the algorithm.

c) Binarization and Colour Inversion

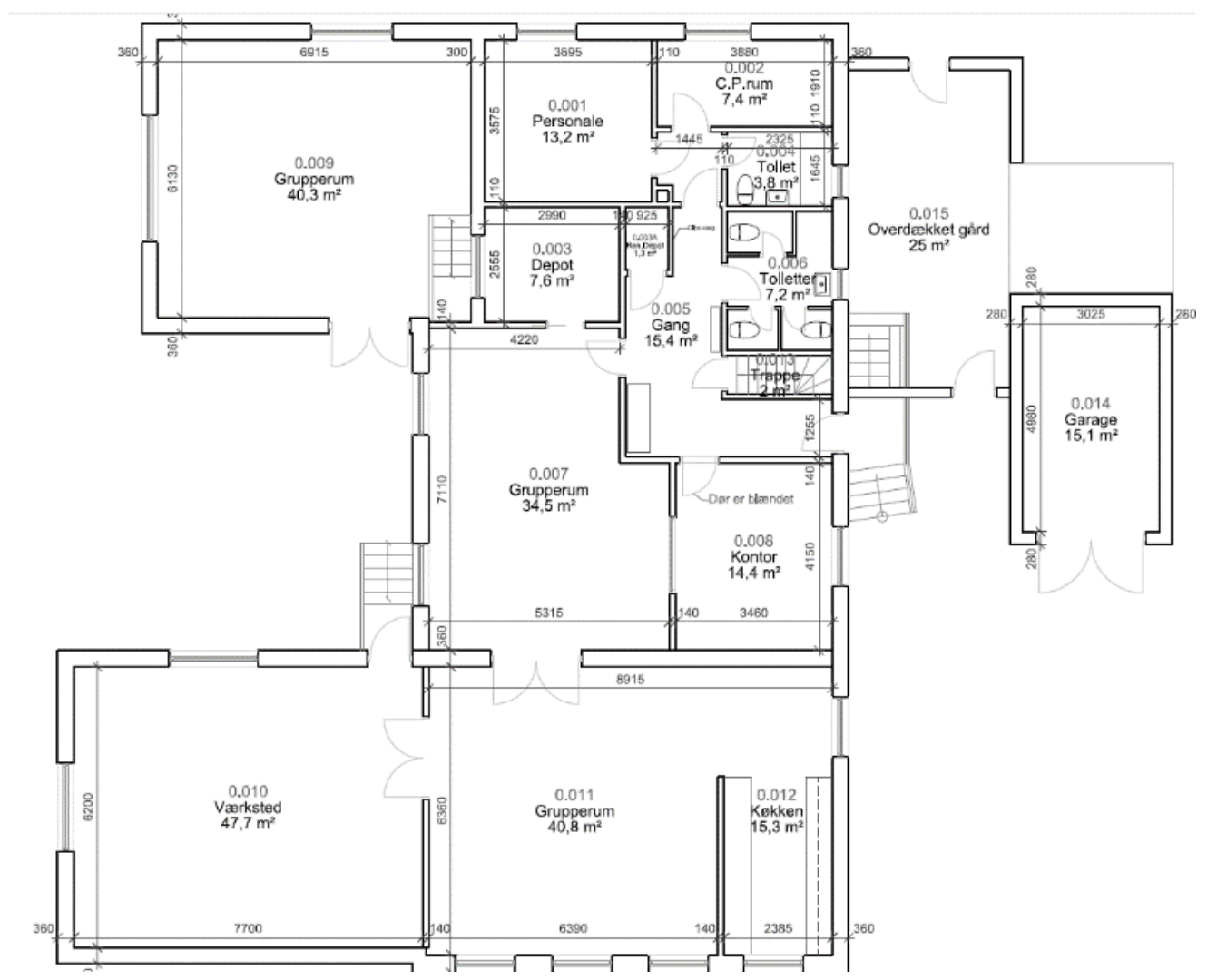

Fig. C5. The binarization of blueprints

Binarization is the method of converting any grey-scale image (multi-tones image) into black–white image (two tone image). The background noise can be greatly avoided via using the binarization. As shown in Fig.C5, the Fig. C4 is binarized and converted to pure white-black image. The binarization is often performed when detecting edges and extracting the object from an image.

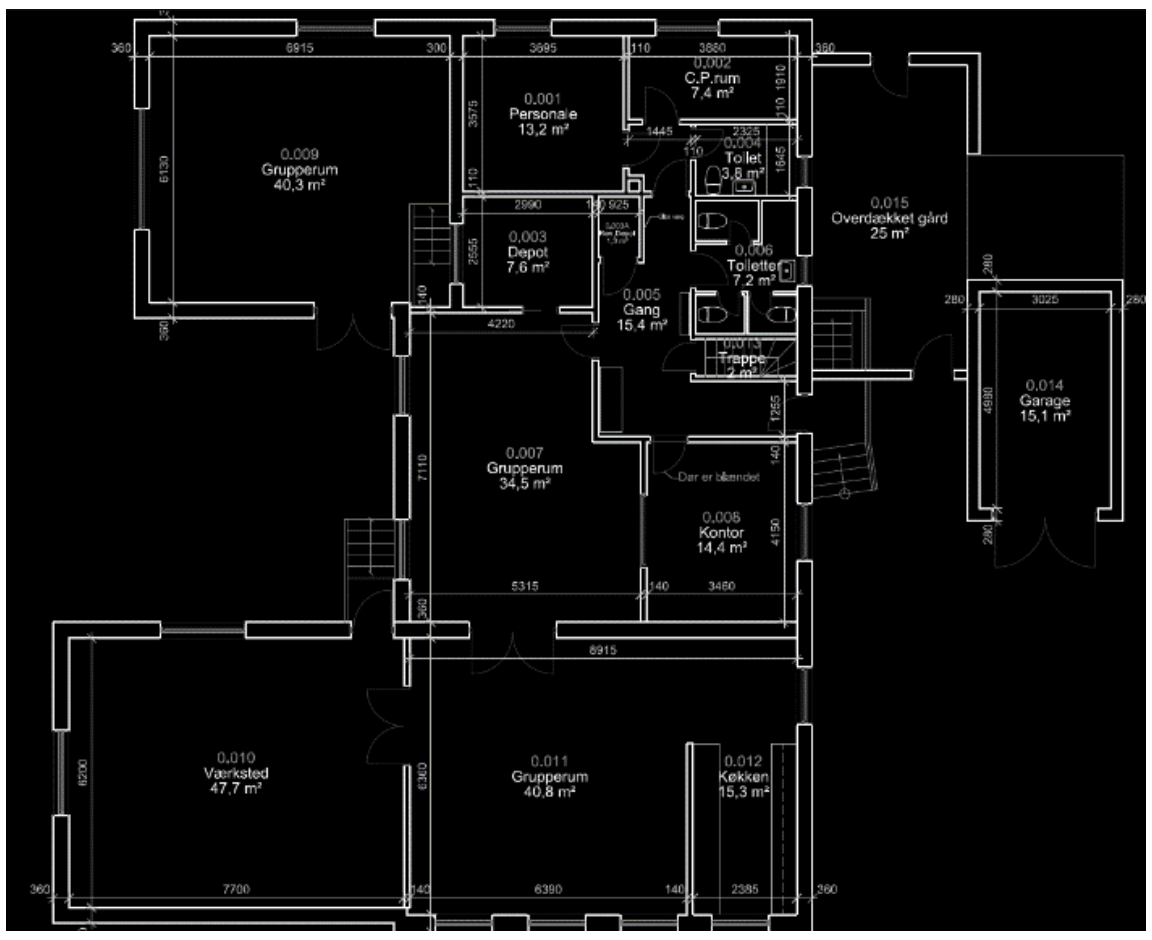

Fig. C6. The color inversion of blueprints

By inverting the binary image in Fig. C5, the pixel values of the image are inverted. From a visual view (as shown in Fig. C6), black pixels will be converted to white, and white pixels will be converted to black. The colour inversion allows easy separation of an object from the dark background.

d) Morphological Operation

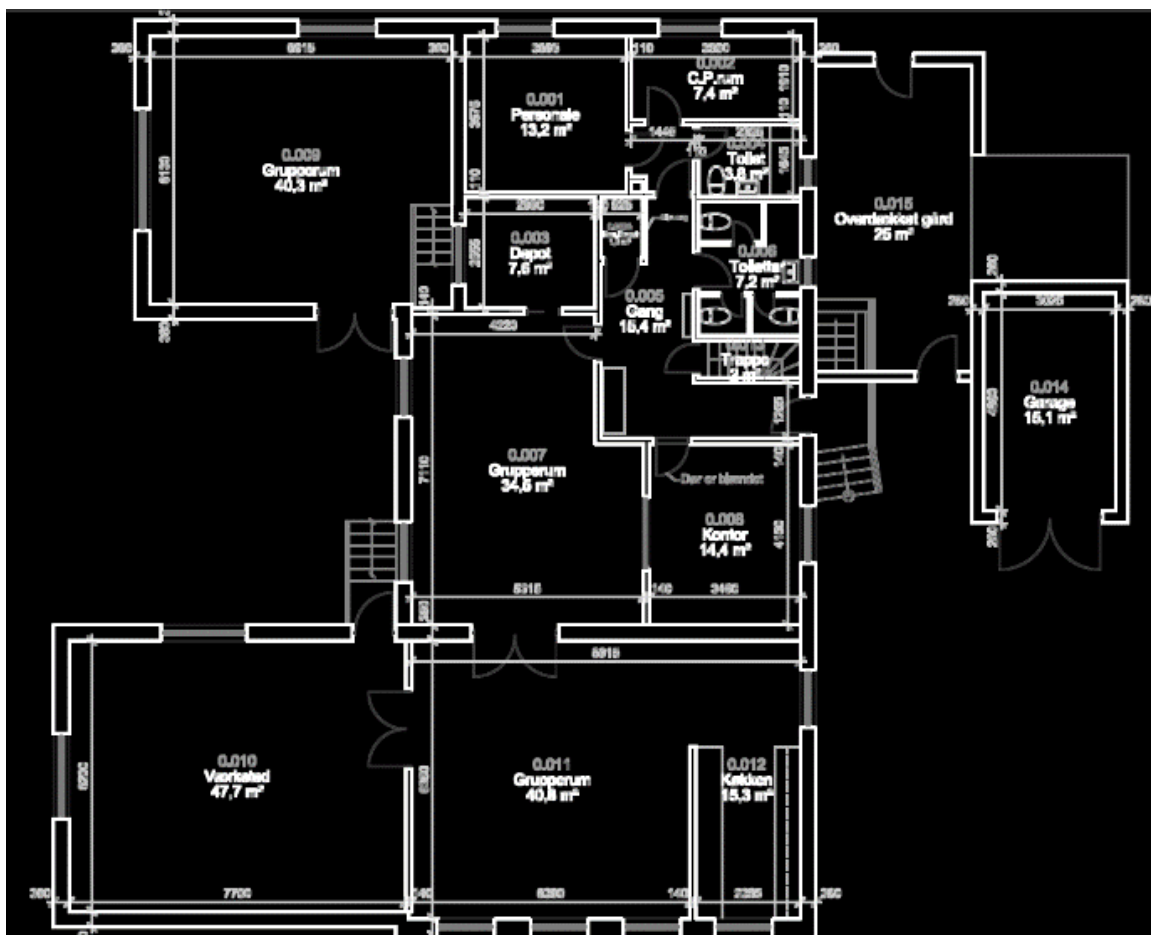

Fig. C7. The morphological operation (e.g., morphological dilation) of blueprints

The morphological operation is also used to reduce noise and improve the clarity in the image processing. Particularly, the morphological processing is normally performed on shapes of features in the binary images, which can add or remove pixels from the images. By this way, the features of interest in an image can be easily extracted.

Generally, the morphological processing includes erosion, dilation, opening and closing. Herein, the dilation (as shown in Fig. C7), which adds pixels to the boundaries of objects in an image, is mostly used in object detection and extraction.  Sometimes, the dilation, which removes pixels from an image, can remove unwanted artifacts and probe the interested shapes contained in the input image. Moreover, morphological closing of an image is a dilation followed by an erosion, which tends to close gaps in the image. Morphological opening is an erosion followed by a dilation, which can polish the edge of large objects.

C3. Contour processing for segmentation

a) Blueprints preparation

In terms of the blueprints preparation, the methods mentioned in the blueprint preprocessing will be used to improve the quality of inputted blueprints in our solution. By the blueprints preparation, we aim to obtain the processed blueprints with less noise and enhanced features, and then improve the accuracy of the segmentation. Next, the boundaries in the blueprints will be detected and extracted for the room segmentation. Finally, the segmented rooms can be coloured randomly.

b) Boundaries detection

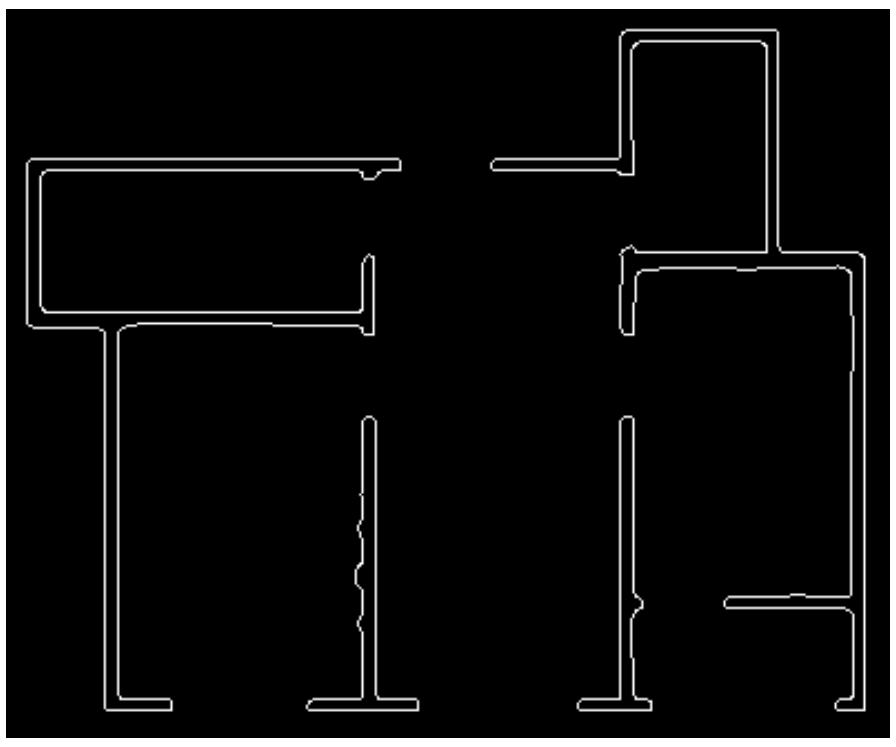

Fig. C8. Preliminary boundaries detection

By using the Canny edge detection function in OpenCV, the main wall in the blueprint can be marked out (see Fig. C8). Canny edge detector is the most popular and most effective way for the edge detection. With the Canny function, the extremely thin and clean edges would be produced. This type of edge detection function consists of noise reduction, gradient calculation, non-maximum suppression, double threshold and edge tracking by hysteresis. The boundaries detection is the preliminary step for the room segmentation.

c) Boundary extraction

In order to divide the whole layout into separate rooms, parallel lines representing main walls need to be calibrated and converted to one line, and then small spaces representing doors and windows need to be connected together. As shown in Fig. C9, the vertical walls in the blueprints are firstly handled. Then, the horizontal walls will be further processed in the same manner.

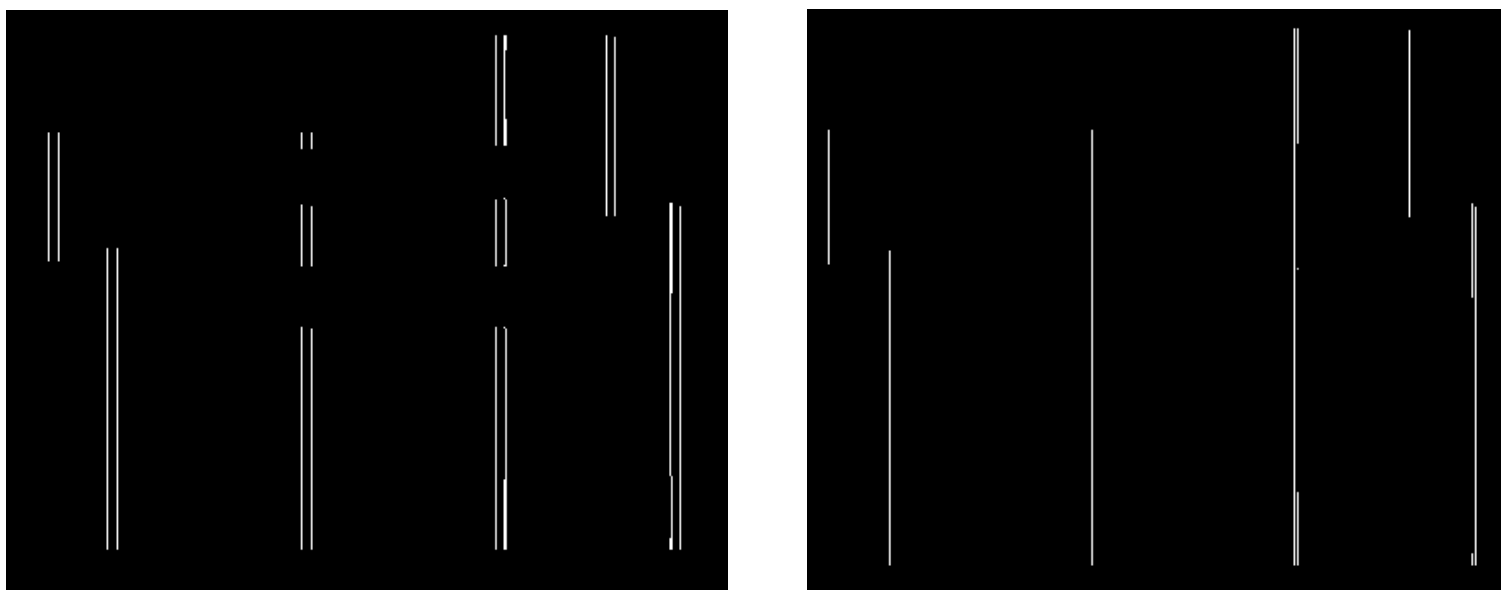

Fig. C9. The boundaries extraction

Via merging the processed vertical and horizontal lines, the whole layout of the blueprints will be clearly segmented into several enclosed rooms.

d) Room colouring

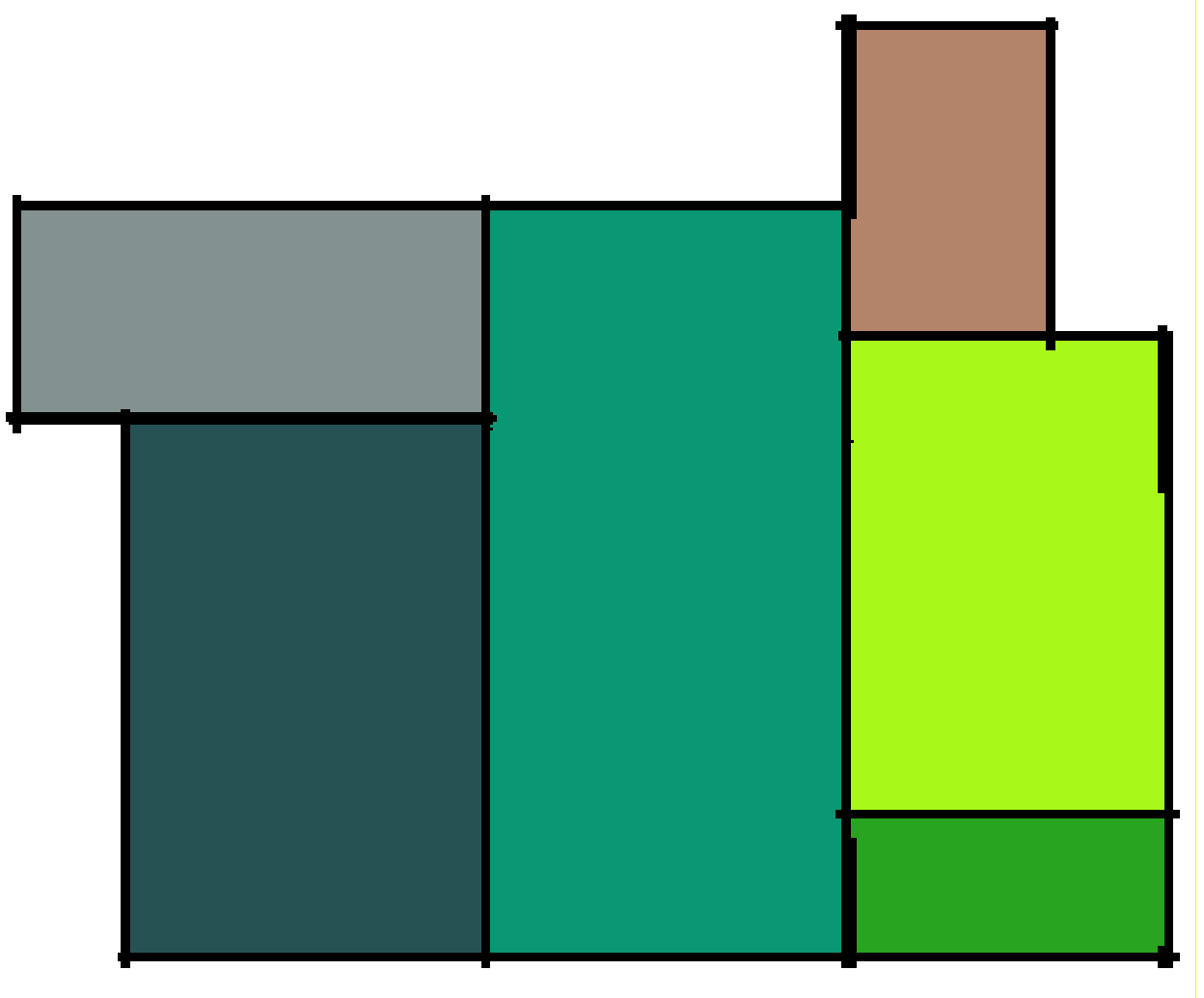

Fig. C10. The room coloring

After the room segmentation above, each enclosed component indicating one room will be numbered and coloured randomly (as shown in Fig.C10). Also, the total pixels in each component can be saved in temporary dataset, which would be applied to the area size calculation below.

C4. Room processing

a) Room counting

The room counting is a following step after the room colouring, which means the component counting. Each component of the blueprint has been segmented and marked out in Section C3. At the same time, the total number of components has been recorded and saved. When counting the room of the blueprint, we just output the total number of all the components. Later, the coloured component referring to each room is overlapped on the original blueprint, as shown in Fig. C11.

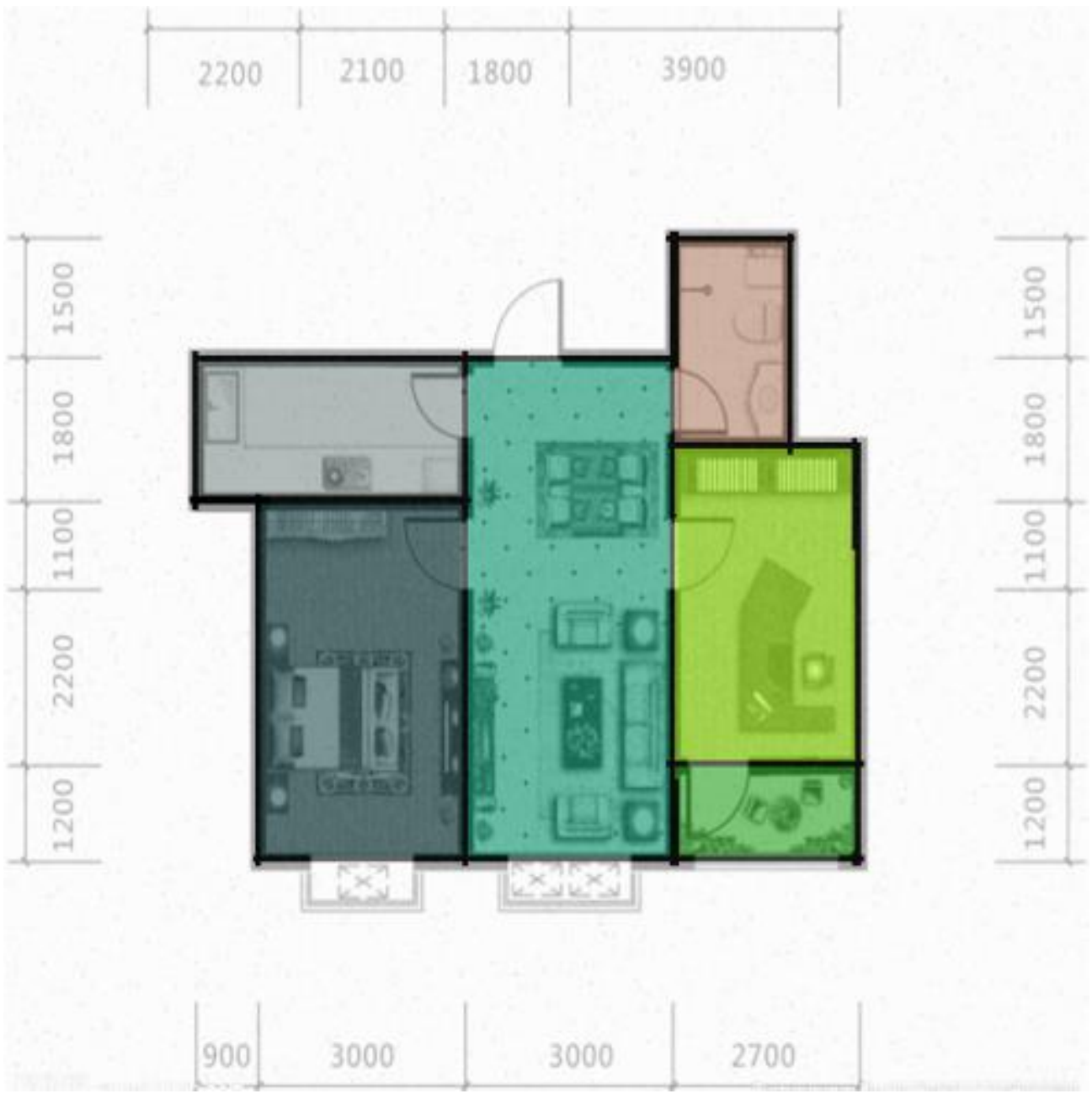

Fig. C11. The room counting

b) Area calculation

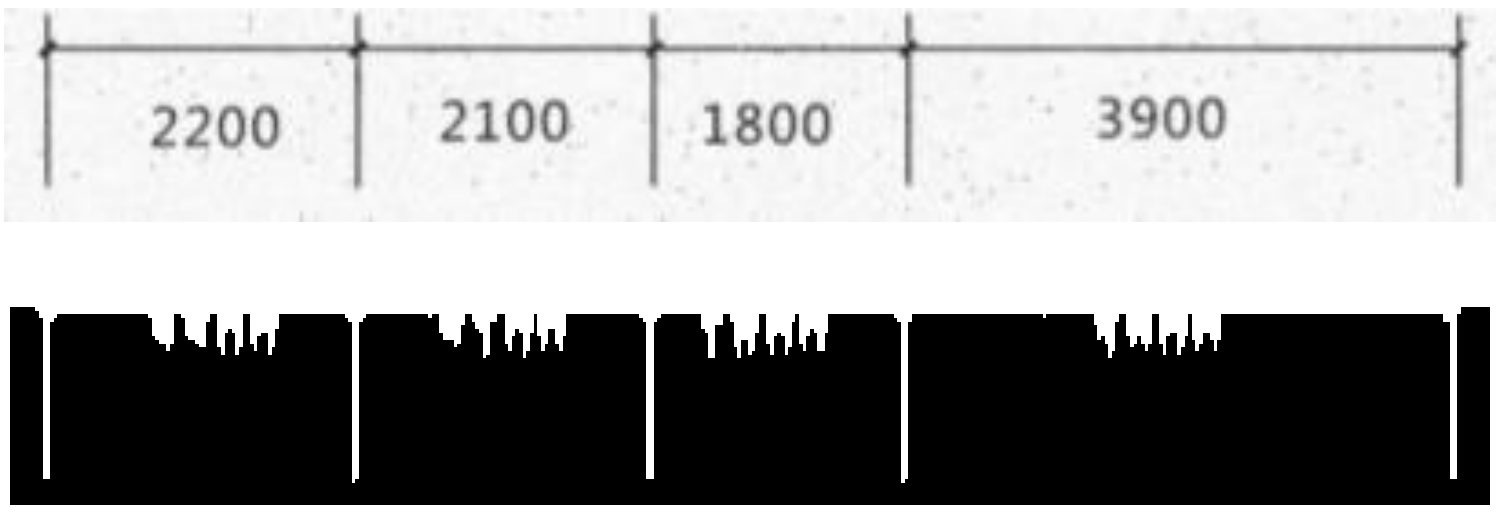

Fig. C12. The ruler detection

Before the calculation of area size, the ruler of each blueprint is supposed to be scanned and confirmed. The area focusing on the ruler area will be firstly circled and extracted, as shown in Fig. C12. Then, the number of the fixed length on the scale ruler will be read. Meanwhile, the volume of pixels corresponding to the fixed length is read. Then, the scale factor can be calculated. The detailed introduction and demonstration of this part can be found in Section C6.(c).1.3.

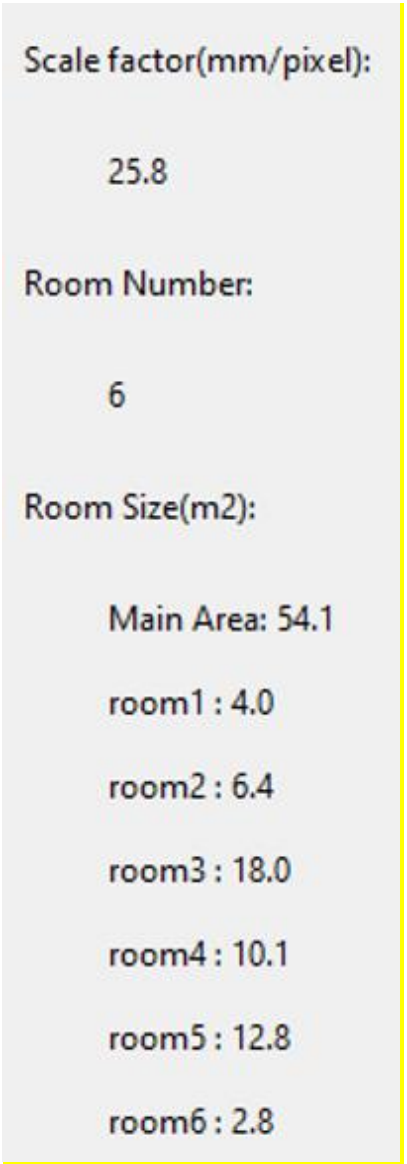

Fig. C13. Results of room area calculation

If the scale factor and the total pixels of each enclosed room are obtained, the size of each room can be simply worked out. As shown in Fig. C13, the area size along with each room is displayed sequentially.

C5. Object Detection

Considering the object detection of blueprints is lack of massive annotated data, template matching can be an alternative technique for finding small parts of an image which are similar to a template image. The template matching in the blueprints interpretation will be much simpler than the AI means (e.g., YOLO5 and SSD) with several benefits, such as no need to label blueprints, accurate bounding boxes and no need for high-performance computation.

Two main components are needed for the template matching, source image and template image. By sliding the patch area in the source image, the patch image will be compared to the template image. The extent to which the patch image is similar to the template image would be calculated by a metric. Herein, the best match can be represented by the maximum value or minimum value of related metrics. In the library of OpenCV, there generally exist six matching methods, such as TM_CCOEFF, TM_CCOEFF_NORMED, TM_CCORR, TM_CCORR_NORMED, TM_SQDIFF and TM_SQDIFF_NORMED.

a) Single Object Matching

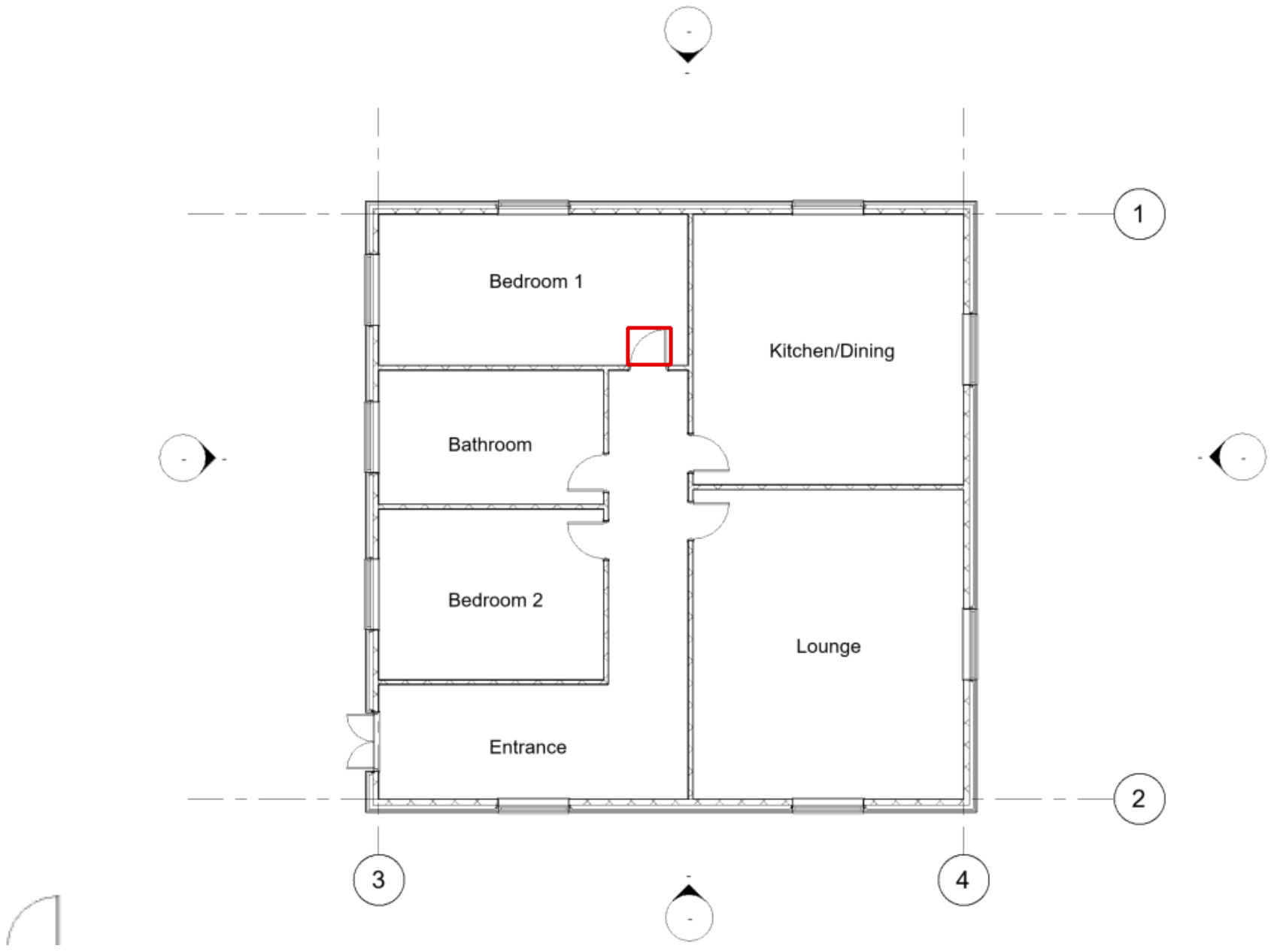

Fig. C14. Single object matching in the blueprint

By detecting the blueprint and parsing the different shapes in the blueprints, the library of different shapes can be saved in the dataset. As shown in Fig. C14, the arch which represents the door can be compared in the library and matched. After pre-processing the image and searching the library, the most similar shape with the best metric in the blueprint can be found and drawn with a red rectangle. However, the same shape would repeat more than one time in the standardized blueprint. The single object matching is hard to meet the requirement in the most of applications. Hence, multi objects matching would be more useful.

b) Multi Objects Matching

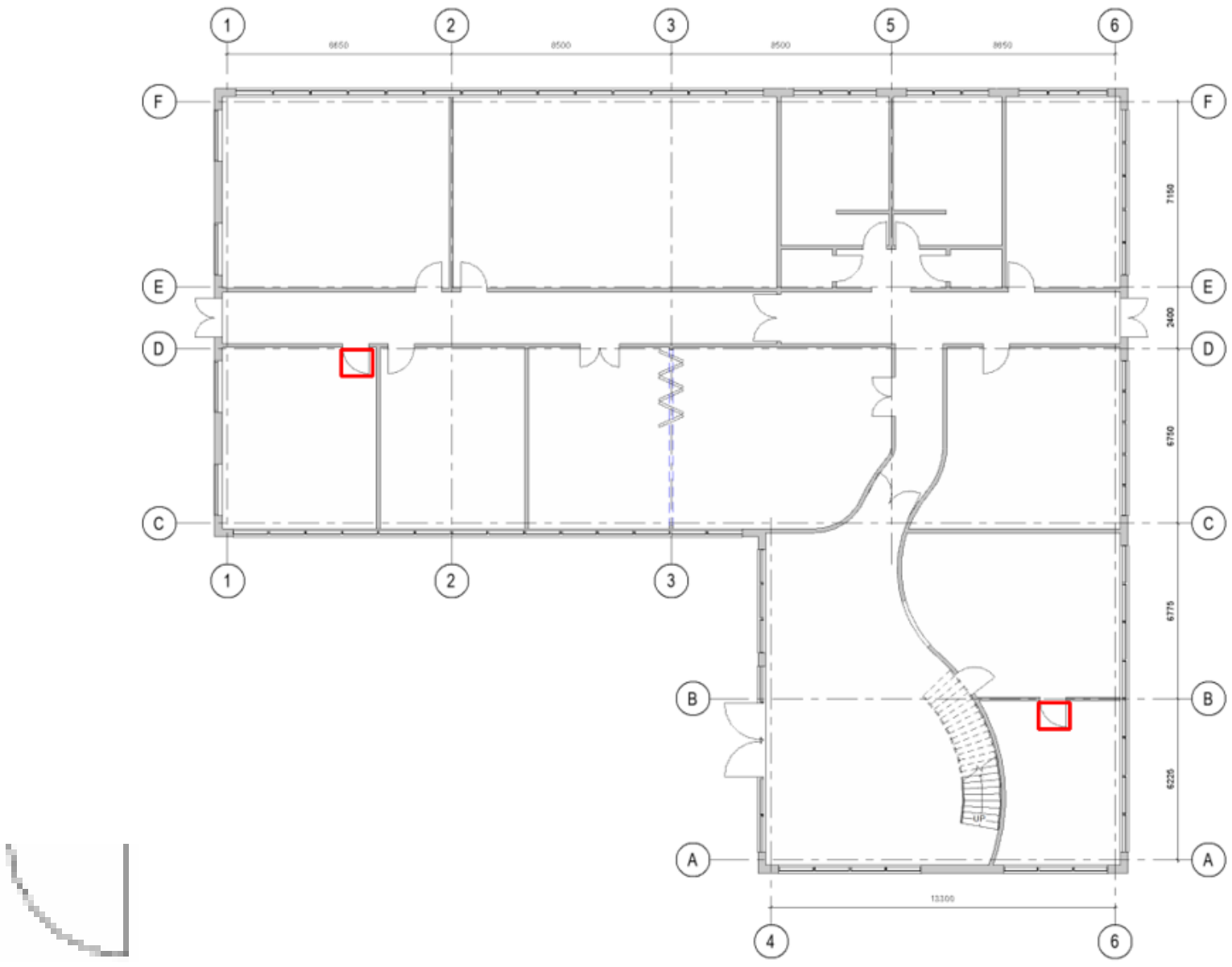

Fig. C15. Multiple objects matching in the blueprint

Compared to the single object matching which only detect one instance of the template, the multiple objects matching which can match similar objects would be more meaningful and often used for the blueprint interpretation (as shown in Fig. C15). By adjusting the template threshold, any object would be found if its value is greater than a fixed threshold. However, the detection accuracy will be influenced when changing the score of threshold.

C6. The usage of OCR in the building blueprints

a) The overview of OCR technology

OCR, is short for Optical Character Recognition, which can be applied to the textual information (e.g., room function, room position and room area) detection and extraction in the image. OCR uses a modular architecture that is open, scalable and workflow controlled. It consists of forms definition, scanning, image pre-processing, text localization, character segmentation, character recognition and post processing. OCR has the ability to turn images of hand-written or printed characters into ASCII data.

b) The experimental setup of OCR

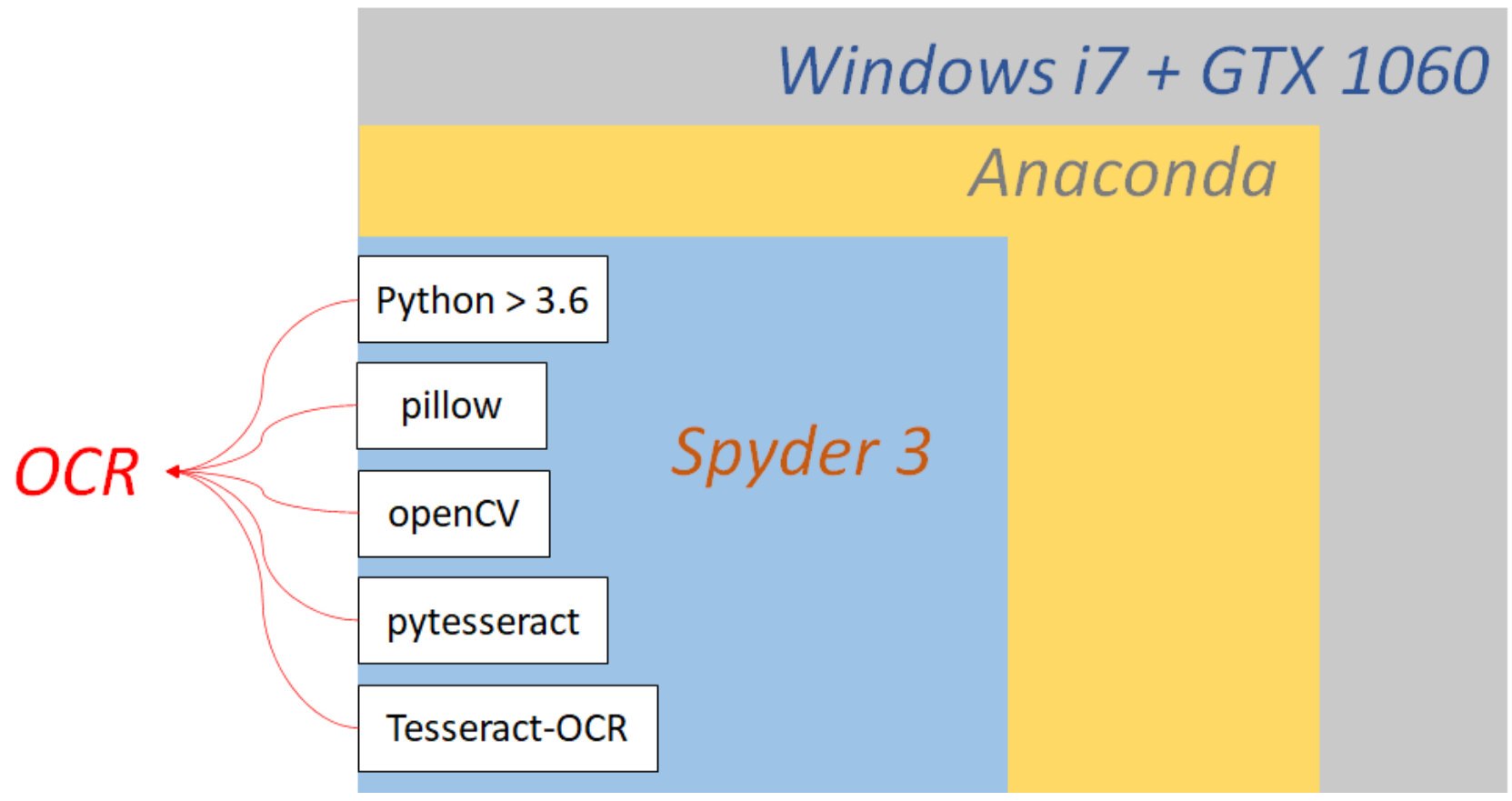

Fig. C4. The experimental implementation of OCR

As shown in Fig. C5, our OCR solution is built on the Anaconda platform with CPU i7 and GPU GTX 1060. And it is written in Python language with several libraries and packages, such as Pillow, OpenCV, Pytesseract and Tesseract-OCR. Herein, Tesseract-OCR is an OCR engine with support for unicode and the ability for 100 languages recognition. Also, Tesseract-OCR can be trained to recognize other types of languages.

c) The use case of OCR in building blueprints

1. Textual information extraction

1.1 Model 1: Basic OCR

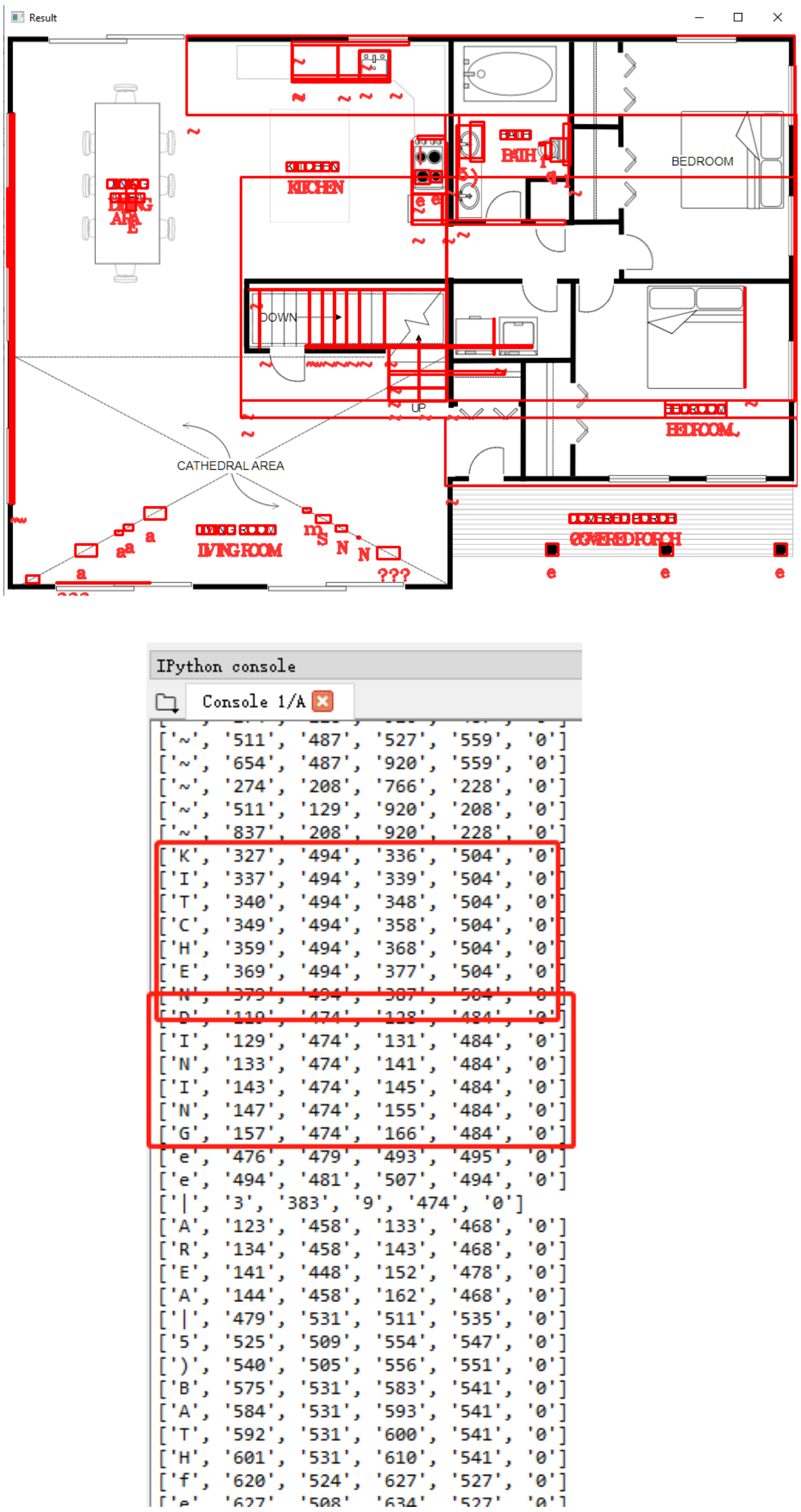

Fig. C6. The experimental result of basic OCR model

In model 1, each character of textual information is detected in red rectangle and printed separately. As shown in Fig. C7, its corresponding position can be displayed and followed by it. However, the semantic information is difficult to extract with separate character detection. Also, the individual position of each character is meaningless.

## 1.2 Model 2: Enhanced OCR

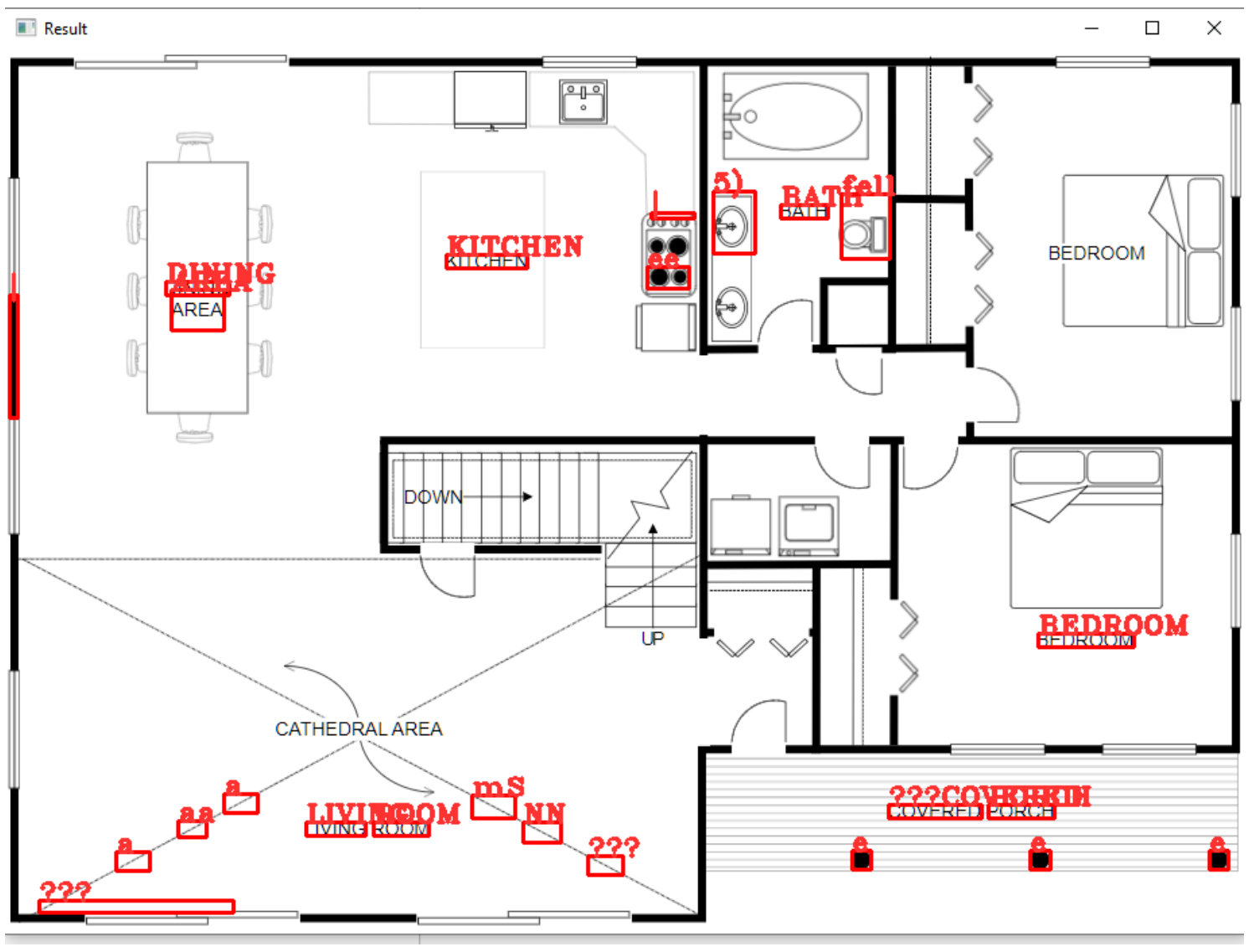

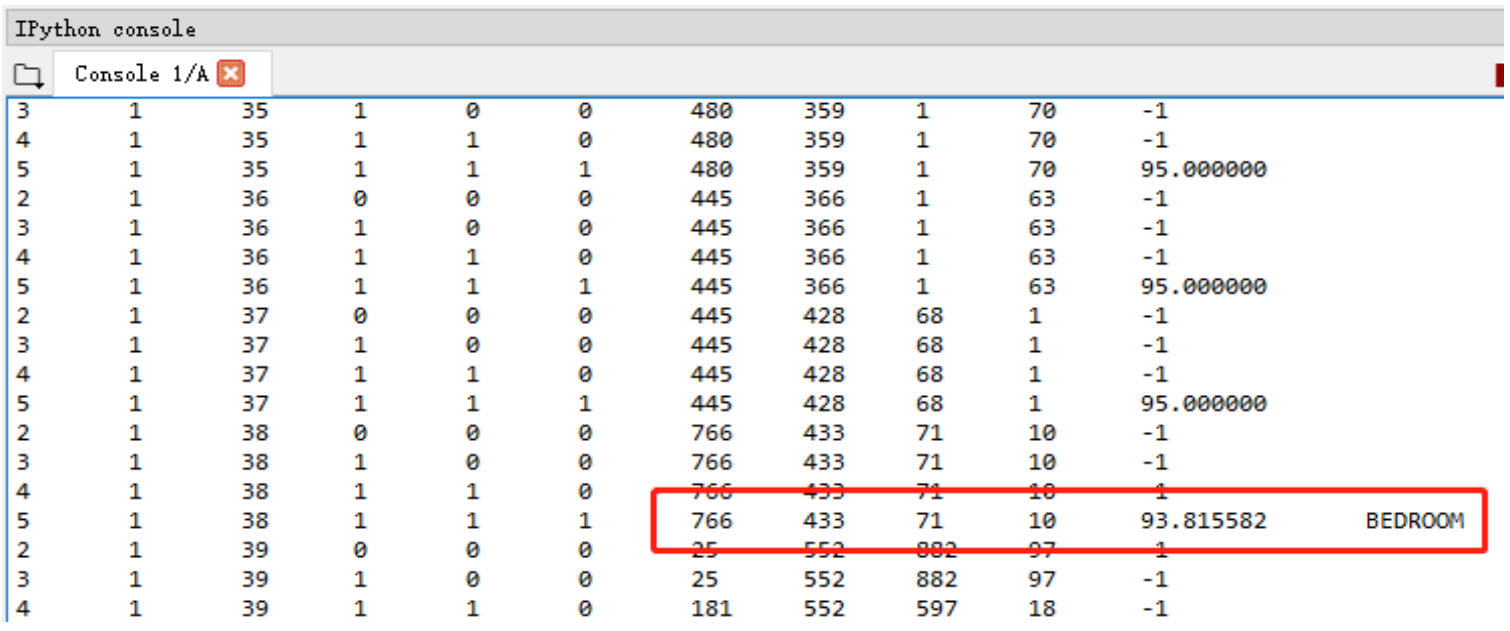

IPython console

Console 1/A

| | | | | | | | | | | | |
|---|---|---|---|---|---|---|---|---|---|---|---|
| 3 | 1 | 35 | 1 | 0 | 0 | 480 | 359 | 1 | 70 | -1 | |
| 4 | 1 | 35 | 1 | 1 | 0 | 480 | 359 | 1 | 70 | -1 | |
| 5 | 1 | 35 | 1 | 1 | 1 | 480 | 359 | 1 | 70 | 95.000000 | |
| 2 | 1 | 36 | 0 | 0 | 0 | 445 | 366 | 1 | 63 | -1 | |
| 3 | 1 | 36 | 1 | 0 | 0 | 445 | 366 | 1 | 63 | -1 | |
| 4 | 1 | 36 | 1 | 1 | 0 | 445 | 366 | 1 | 63 | -1 | |
| 5 | 1 | 36 | 1 | 1 | 1 | 445 | 366 | 1 | 63 | 95.000000 | |
| 2 | 1 | 37 | 0 | 0 | 0 | 445 | 428 | 68 | 1 | -1 | |
| 3 | 1 | 37 | 1 | 0 | 0 | 445 | 428 | 68 | 1 | -1 | |
| 4 | 1 | 37 | 1 | 1 | 0 | 445 | 428 | 68 | 1 | -1 | |
| 5 | 1 | 37 | 1 | 1 | 1 | 445 | 428 | 68 | 1 | 95.000000 | |
| 2 | 1 | 38 | 0 | 0 | 0 | 766 | 433 | 71 | 10 | -1 | |
| 3 | 1 | 38 | 1 | 0 | 0 | 766 | 433 | 71 | 10 | -1 | |
| 4 | 1 | 38 | 1 | 1 | 0 | 766 | 433 | 71 | 10 | -1 | |
| 5 | 1 | 38 | 1 | 1 | 1 | 766 | 433 | 71 | 10 | 93.815582 | BEDROOM |
| 2 | 1 | 39 | 0 | 0 | 0 | 25 | 552 | 882 | 97 | -1 | |
| 3 | 1 | 39 | 1 | 0 | 0 | 25 | 552 | 882 | 97 | -1 | |
| 4 | 1 | 39 | 1 | 1 | 0 | 181 | 552 | 597 | 18 | -1 | |

Fig. C8. The experimental result of enhanced OCR model

With the enhanced OCR model, textual information can be read and outputted consecutively and completely. Its semantic information is more readable and valuable. Furthermore, the relative position of characters in the blueprint can be measured and displayed. As shown in Fig. C9, “BEDROOM” with coordinates can represent the function of the room and its relative position in the blueprint. The enhanced OCR model would be commonly used in the blueprint interpretation for textual information extraction.

## 1.3 Area size calculation

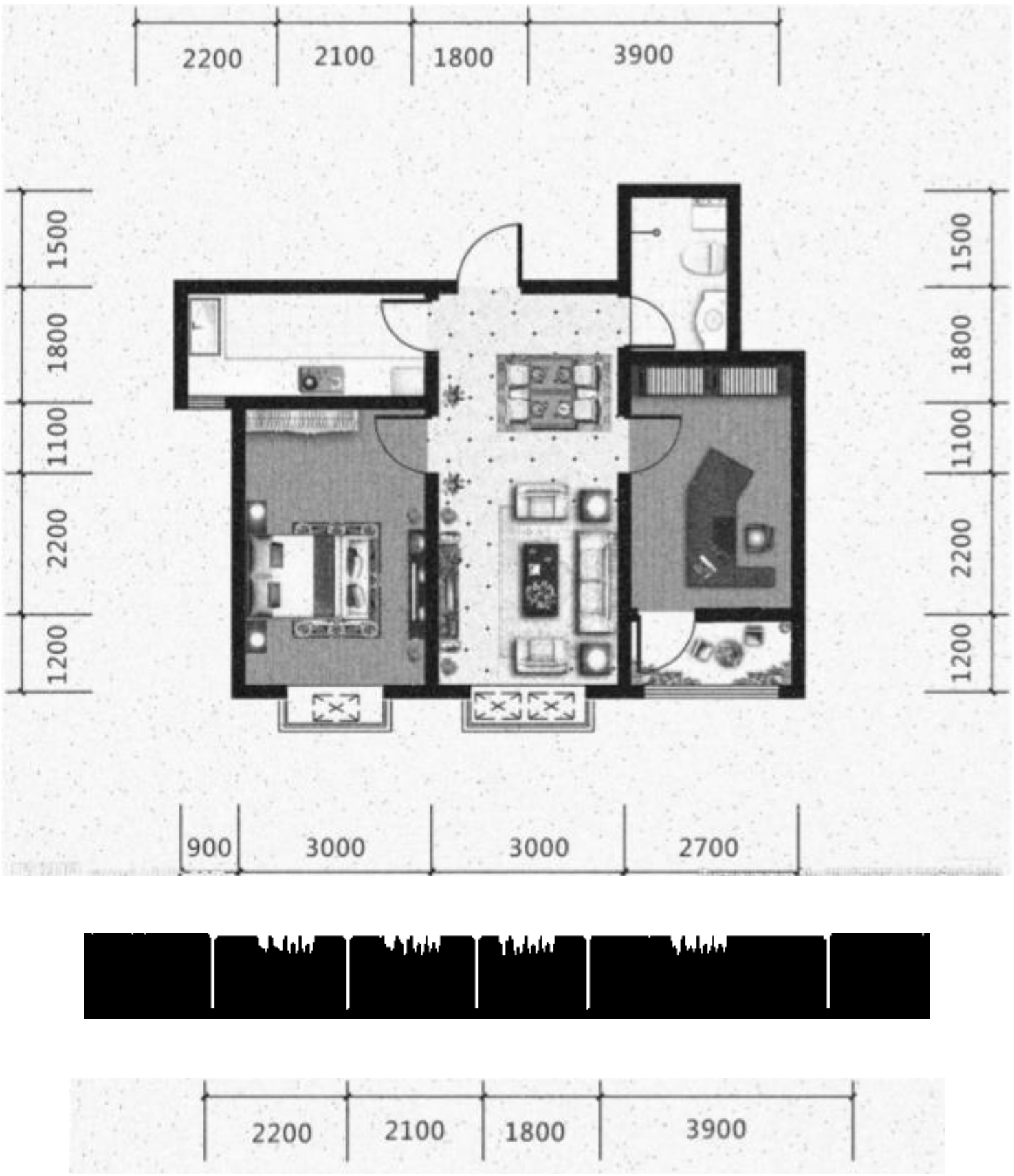

Fig. C19. The ruler detection

By splitting RGB tones of the blueprint, the mask image of the blueprint can be easily displayed. Then, the Contrast Limited Adaptive Histogram Equalization (CLAHE) will be used to reduce the noise in the blueprint and brighten the ruler scaler (with less information). Next, the bright part would be cropped in a small rectangle, as shown in Fig. C19. With OCR technology, the value of four sectors on the ruler can be represented. Also, the corresponding pixels on each sector can be read. Finally, the number of pixels will be matched with the value of each sector on the ruler (i.e., scale factor).

After confirming the contour of each room, the total of pixels on this area can be detected and obtained, and then the scale factor will be used to calculate the area size.

1.4 Area-size extraction and comparation

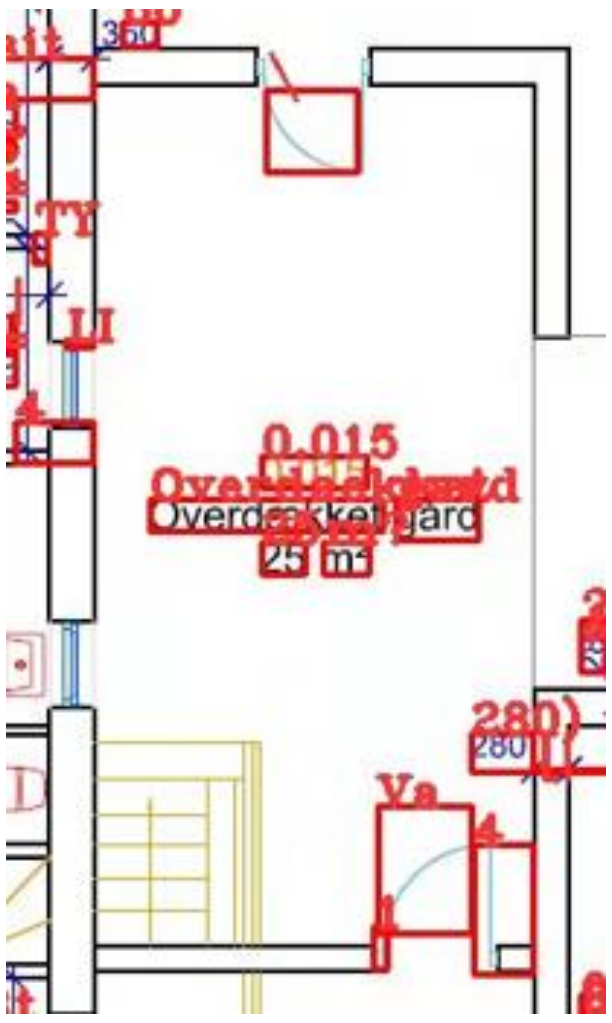

```
['5', '1', '98', '1', '12', '2', '939', '311', '102', '14', '87.098846', 'Overdaekket']
['5', '1', '98', '1', '12', '3', '1047', '311', '36', '18', '96.845726', 'gard']
['4', '1', '98', '1', '13', '0', '486', '330', '549', '21', '-1']
['5', '1', '98', '1', '13', '1', '486', '344', '11', '7', '57.102768', '|']
['5', '1', '98', '1', '13', '2', '537', '335', '46', '14', '55.064919', '0.003']
['5', '1', '98', '1', '13', '3', '988', '330', '19', '14', '91.014313', '25']
['5', '1', '98', '1', '13', '4', '1015', '330', '20', '14', '91.014313', 'm?']
```

Fig. C10. The area size extraction

As shown in Fig. C11, the function of detected room and its corresponding size can be outputted along. Compared with the area size calculated by the method in Section 4.3.2, they can be mutually verified. This way would improve the accuracy of blueprints interpretation.

## C7. The unified platform for the blueprint interpretation

By combing a series of technologies above, a unified platform is further implemented to interpret the comprehensive information of one blueprint, as shown in Fig. C12. After contour processing, the room can be segmented and counted. By reading the ruler and counting the pixel in the blueprint, the area of each space can be calculated. Meanwhile, with the fine-grained blueprints, the area of each space can be extracted by OCR technology. Thus, these two ways can be mutually verified.

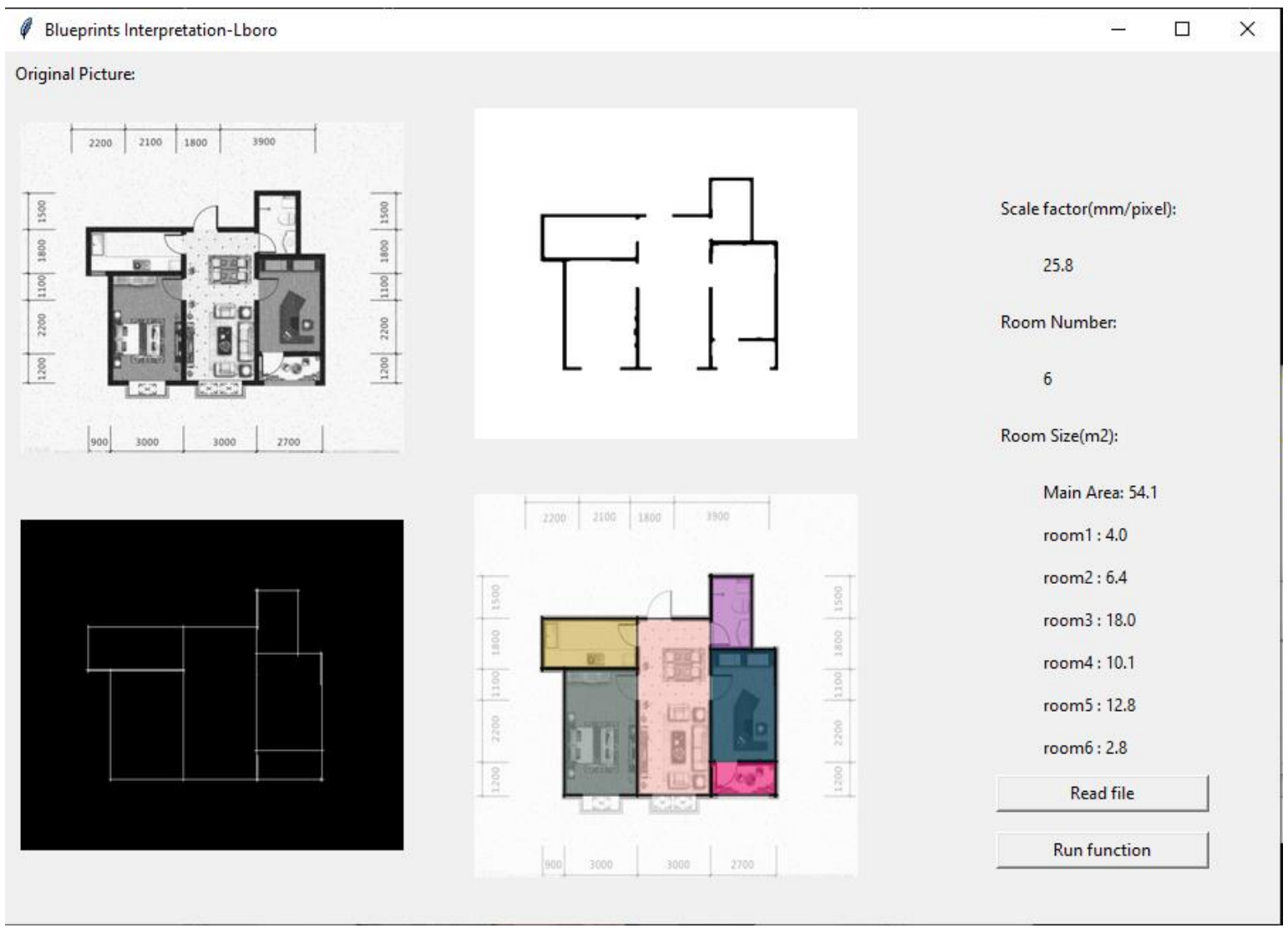

Fig. C13. The unified platform for blueprints interpretation

# References

Advisen Ltd. (2013). Focusing on Property Risk Management. Retrieved from https://erm.ncsu.edu/library/article/survey-property-risk-management

Ahmed, S., Liwicki, M., Weber, M., & Dengel, A. (2011). Improved automatic analysis of architectural floor plans. *Proceedings of the International Conference on Document Analysis and Recognition, ICDAR*, 864–869. https://doi.org/10.1109/ICDAR.2011.177

Bulat, A., & Tzimiropoulos, G. (2016). Human Pose Estimation via Convolutional Part Heatmap Regression. In *Leibe B., Matas J., Sebe N., Welling M. (eds) Computer Vision – ECCV 2016. ECCV 2016. Lecture Notes in Computer Science, vol 9911.* (pp. 717–732). Springer, Cham. https://doi.org/10.1007/978-3-319-46478-7_44

Dosch, P., Tombre, K., Ah-Soon, C., & Masini, G. (2000). A complete system for the analysis of architectural drawings. *International Journal on Document Analysis and Recognition*, *3*(2), 102–116. https://doi.org/10.1007/PL00010901

Figueroa, M. (2013). Fire sprinklers: Protecting lives, property and assisting water conservation efforts. *Journal-American Water Works Association*, *105*(1), E11–E18.

Frank, K., Gravestock, N., Spearpoint, M., & Fleischmann, C. (2013). A review of sprinkler system effectiveness studies. *Fire Science Reviews*, *2*(1), 1–19.

Gimenez, L., Hippolyte, J.-L., Robert, S., Suard, F., & Zreik, K. (2015). Review: reconstruction of 3D building information models from 2D scanned plans. *Journal of Building Engineering*, *2*, 24–35. https://doi.org/10.1016/j.jobe.2015.04.002

He, K., Zhang, X., Ren, S., & Sun, J. (2016). Deep Residual Learning for Image Recognition. In *2016 IEEE Conference on Computer Vision and Pattern Recognition (CVPR)* (Vol. 37, pp. 770–778). IEEE. https://doi.org/10.1109/CVPR.2016.90

Holborn, P. G., Nolan, P. F., & Golt, J. (2004). An analysis of fire sizes, fire growth rates and times between events using data from fire investigations. *Fire Safety Journal*, *39*(6), 481–524.

Huang, W., & Zheng, H. (2018). Architectural drawings recognition and generation through machine learning. *Recalibration on Imprecision and Infidelity - Proceedings of the 38th Annual Conference of the Association for Computer Aided Design in Architecture, ACADIA 2018*, 156–165.

Jang, H., Yu, K., & Yang, J. H. (2020). Indoor reconstruction from floorplan images with a deep learning approach. *ISPRS International Journal of Geo-Information*, *9*(2). https://doi.org/10.3390/ijgi9020065

Kalervo, A., Ylioinas, J., Häikiö, M., Karhu, A., & Kannala, J. (2019). CubiCasa5K: A Dataset and an Improved Multi-task Model for Floorplan Image Analysis. *Lecture Notes in Computer Science (Including Subseries Lecture Notes in Artificial Intelligence and Lecture Notes in Bioinformatics)*, *11482 LNCS*, 28–40. https://doi.org/10.1007/978-3-030-20205-7_3

Kim, S., & Park, S. (2018). Application of Style Transfer in the Vectorization Process of Floorplans. In *10th International Conference on Geographic Information Science (GIScience 2018).* (pp. 1–6). Schloss Dagstuhl-Leibniz-Zentrum fuer Informatik. https://doi.org/10.4230/LIPIcs.GIScience.2018.39

Liu, C., Wu, J., Kohli, P., & Furukawa, Y. (2017). Raster-to-Vector: Revisiting Floorplan Transformation. In *2017 IEEE International Conference on Computer Vision (ICCV)* (Vol. 2017-Octob, pp. 2214–2222). IEEE. https://doi.org/10.1109/ICCV.2017.241

Lladós, J., López-Krahe, J., & Martí, E. (1997). A system to understand hand-drawn floor plans using subgraph isomorphism and Hough transform. *Machine Vision and Applications*, *10*(3), 150–158. https://doi.org/10.1007/s001380050068

Lu, T., Yang, H., Yang, R., & Cai, S. (2007). Automatic analysis and integration of architectural drawings. *International Journal on Document Analysis and Recognition*, *9*(1), 31–47. https://doi.org/10.1007/s10032-006-0029-6

Paudel, A., Dhakal, R., & Bhattarai, S. (2021). Room Classification on Floor Plan Graphs using Graph

Neural Networks. *Proceedings of ACM Conference (Conference'17)*, *1*(1).
Rica, E., Moreno-García, C. F., Álvarez, S., & Serratosa, F. (2020). Reducing human effort in engineering drawing validation. *Computers in Industry*, *117*, 103198.
Ryall, K., Shieber, S., Marks, J., & Mazer, M. (1995). Semi-automatic delineation of regions in floor plans. *Proceedings of the International Conference on Document Analysis and Recognition, ICDAR*, *2*, 964–969. https://doi.org/10.1109/ICDAR.1995.602062
Sharma, D., Gupta, N., Chattopadhyay, C., & Mehta, S. (2019). A novel feature transform framework using deep neural network for multimodal floor plan retrieval. *International Journal on Document Analysis and Recognition*, *22*(4), 417–429. https://doi.org/10.1007/s10032-019-00340-1
Siu-hang, Wong, K. H., Yu, Y. K., & Chang, M. Y. (2009). Abstract Highly Automatic Approach to Architectural Floorplan Image Understanding & Model Generation. *Pattern Recognition*.
Sølvsten, S. (2022). *The economic net benefits of loss prevention technologies in the context of risk management and insurance*. University of Southern Denmark.
Yamada, M., Wang, X., & Yamasaki, T. (2021). Graph Structure Extraction from Floor Plan Images and Its Application to Similar Property Retrieval. *Digest of Technical Papers - IEEE International Conference on Consumer Electronics*, *2021-Janua*. https://doi.org/10.1109/ICCE50685.2021.9427580
Zeng, Z., Li, X., Yu, Y. K., & Fu, C.-W. (2019). Deep Floor Plan Recognition Using a Multi-Task Network With Room-Boundary-Guided Attention. In *2019 IEEE/CVF International Conference on Computer Vision (ICCV)* (Vol. 2019-Octob, pp. 9095–9103). IEEE. https://doi.org/10.1109/ICCV.2019.00919
Zhao, Y., Deng, X., & Lai, H. (2020). A Deep Learning-Based Method to Detect Components from Scanned Structural Drawings for Reconstructing 3D Models. *Applied Sciences*, *10*(6), 2066. https://doi.org/10.3390/app10062066